\documentclass[final]{cvpr}

\usepackage{times}
\usepackage{epsfig}
\usepackage{graphicx}
\usepackage{amsmath}
\usepackage{amssymb}

\usepackage{enumitem} 
\usepackage{caption}
\usepackage{subcaption}
\usepackage{tabularx, booktabs} 
\makeatletter
\@namedef{ver@eso-pic.sty}{}
\makeatother
\usepackage{pdfpages}

\usepackage[pagebackref=true,breaklinks=true,colorlinks,bookmarks=false]{hyperref}

\def\cvprPaperID{4630} 
\def\confYear{CVPR 2021}

\begin{document}

\title{Background-Aware Pooling and Noise-Aware Loss for Weakly-Supervised Semantic Segmentation}

\author{Youngmin Oh
\quad\quad\quad
Beomjun Kim
\quad\quad\quad
Bumsub Ham\thanks{Corresponding author.}
\\
School of Electrical and Electronic Engineering, Yonsei University
}
\maketitle
\thispagestyle{empty}

\begin{abstract}
We address the problem of weakly-supervised semantic segmentation~(WSSS) using bounding box annotations. Although object bounding boxes are good indicators to segment corresponding objects, they do not specify object boundaries, making it hard to train convolutional neural networks~(CNNs) for semantic segmentation. We find that background regions are perceptually consistent in part within an image, and this can be leveraged to discriminate foreground and background regions inside object bounding boxes. To implement this idea, we propose a novel pooling method, dubbed background-aware pooling~(BAP), that focuses more on aggregating foreground features inside the bounding boxes using attention maps. This allows to extract high-quality pseudo segmentation labels to train CNNs for semantic segmentation, but the labels still contain noise especially at object boundaries. To address this problem, we also introduce a noise-aware loss (NAL) that makes the networks less susceptible to incorrect labels. Experimental results demonstrate that learning with our pseudo labels already outperforms state-of-the-art weakly- and semi-supervised methods on the PASCAL VOC 2012 dataset, and the NAL further boosts the performance.
\end{abstract}

\section{Introduction}
\begin{figure}[t]
\captionsetup[subfigure]{labelformat=empty}
\tiny
\centering
\begin{subfigure}{.117\textwidth}
  \centering
  \frame{\includegraphics[width=\textwidth]{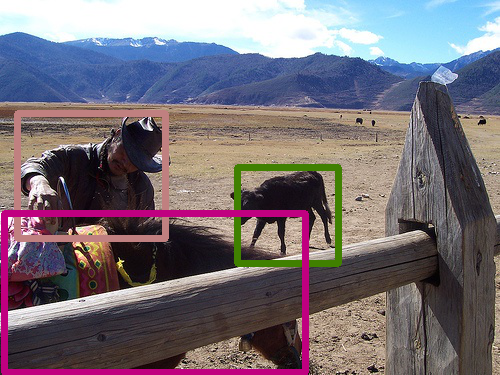}}\vspace{-0.1cm}
  \caption{Input image.}
\end{subfigure}
\begin{subfigure}{.117\textwidth}
  \centering
  \frame{\includegraphics[width=\textwidth]{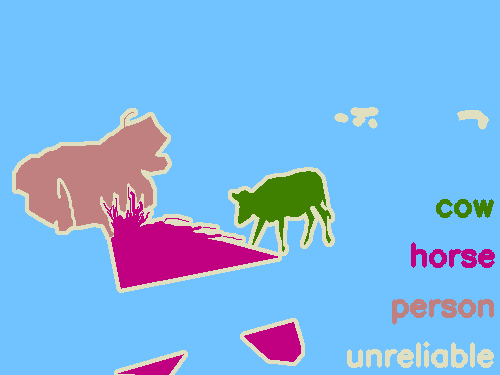}}\vspace{-0.1cm}
  \caption{Ground truth.}
\end{subfigure}
\begin{subfigure}{.117\textwidth}
  \centering
  \frame{\includegraphics[width=\textwidth]{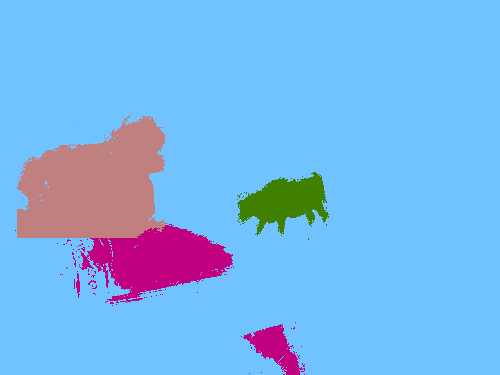}}\vspace{-0.1cm}
  \caption{Ours.}
\end{subfigure}
\begin{subfigure}{.117\textwidth}
  \centering
  \frame{\includegraphics[width=\textwidth]{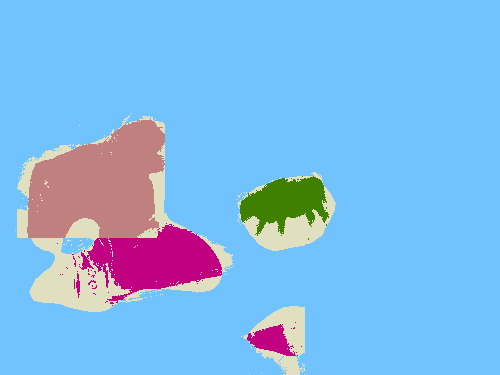}}\vspace{-0.1cm}
  \caption{Ours$^*$.}
\end{subfigure}

\begin{subfigure}{.117\textwidth}
  \centering
  \frame{\includegraphics[width=\textwidth]{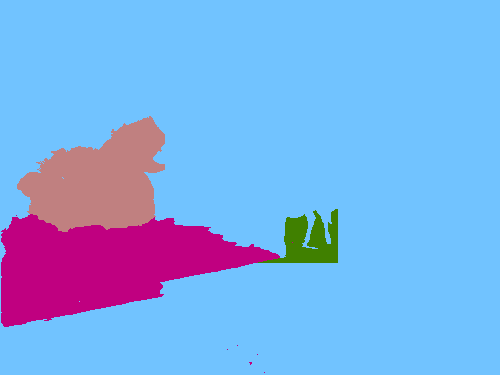}}\vspace{-0.1cm}
  \caption{GrabCut~\cite{GRABCUT04Rother}.}
\end{subfigure}
\begin{subfigure}{.117\textwidth}
  \centering
  \frame{\includegraphics[width=\textwidth]{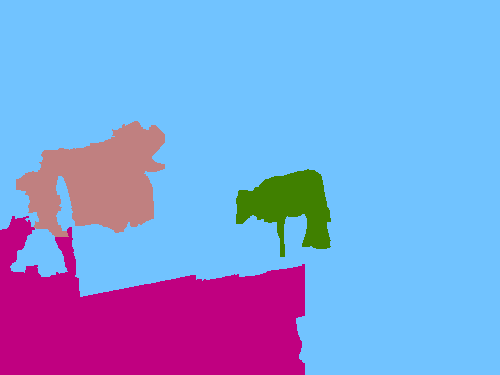}}\vspace{-0.1cm}
  \caption{MCG\footnotemark~\cite{MCG16Pont}.}
\end{subfigure}
\begin{subfigure}{.117\textwidth}
  \centering
  \frame{\includegraphics[width=\textwidth]{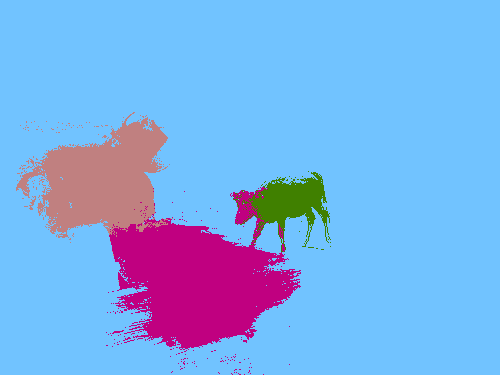}}\vspace{-0.1cm}
  \caption{WSSL~\cite{WSSL15Pap}.}
\end{subfigure}
\begin{subfigure}{.117\textwidth}
  \centering
  \frame{\includegraphics[width=\textwidth]{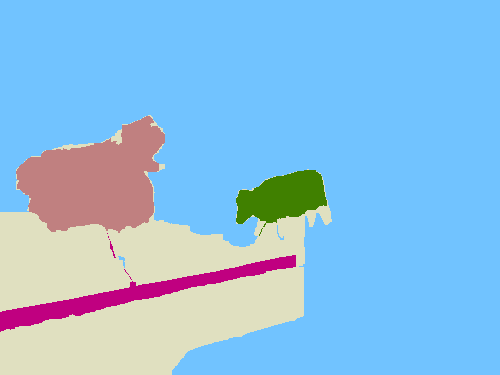}}\vspace{-0.1cm}
  \caption{SDI~\cite{SDI17Kho}.}
\end{subfigure}
\vspace{-.25cm}
\caption{\small{Visual comparison of pseudo ground-truth labels. Our approach generates better segmentation labels than other WSSS methods using object bounding boxes~(WSSL~\cite{WSSL15Pap} and SDI~\cite{SDI17Kho}). Hand-crafted methods~(GrabCut~\cite{GRABCUT04Rother} and MCG~\cite{MCG16Pont}) fail to segment object boundaries. Ours$^*$: Ours with an indication of unreliable regions. Best viewed in color.}}
\vspace{-.6cm}
\label{fig:teaser}
\end{figure}

\footnotetext{For MCG, we compute intersection-over-union~(IoU) scores using pairs of segment proposals and bounding boxes, and choose the best one for each box.}

Semantic segmentation is one of the fundamental tasks in computer vision, and has received a lot of attention over the last decades. It aims at assigning a semantic label to each pixel, which can be leveraged to various applications including scene understanding, autonomous driving, image editing, and robotics. Supervised methods based on convolutional neural networks~(CNNs)~\cite{DEEPLAB14Chen,FCN15Long,UNET15Ronn} have achieved remarkable success in semantic segmentation, but they require lots of training samples with pixel-level labels, which are extremely labor-intensive to annotate, to train networks. Weakly-supervised semantic segmentation~(WSSS) has recently been introduced to exploit a weak form of supervisory signals such as image-level labels~\cite{ICD20Fan,ASSOCIATE18Fan,SeeNet18Hou,SEC16Kol,FICKLE19Lee,REVISIT18Wei,RDM19Zhang}, points~\cite{POINT16Bearman}, scribbles~\cite{SCRIBBLE16Lin,SCRIBBLEREGUL18Tang,SCRIBBLERW17Vern}, and object bounding boxes~\cite{BOXSUP15Dai,SDI17Kho,WSSL15Pap,BOXDRIVEN19Song}. WSSS methods using image-level labels typically leverage class activation maps~(CAMs)~\cite{CAM16Zhou}, obtained from CNNs for image classification using global average pooling~(GAP), to localize objects. Since CAMs tend to highlight discriminative parts, these methods more or less resort to off-the-shelf saliency detectors~\cite{S4NET19Fan,DSOD17Hou,SOD13Jiang,ISOD17Li,SESOD17Xiao}. This, however, requires additional pixel-level ground-truth annotations for salient objects. Other approaches attempt to exploit object bounding boxes. They are easy to annotate compared to pixel-level labels and provide rich semantics to localize objects. The object bounding boxes, however, contain a mixture of foreground and background, and do not specify exquisite object boundaries. To overcome this, recent approaches~\cite{BOXSUP15Dai,SDI17Kho,BTS20Kul} use off-the-shelf segmentation methods~\cite{MCG16Pont,GRABCUT04Rother}.

We introduce a simple yet effective WSSS method using bounding box annotations. In particular, we investigate two aspects of this problem -- How can we generate high-quality but possibly noisy pixel-level labels~(\emph{i.e.},~a pseudo ground truth) from object bounding boxes~(Fig.~\ref{fig:teaser})? How can we train CNNs for semantic segmentation~(\emph{e.g.}, DeepLab~\cite{DEEPLAB14Chen,DEEPLAB18Chen}) with noisy segmentation labels? Motivated by the methods using image-level labels~\cite{AFFINITY18Ahn,SeeNet18Hou,FICKLE19Lee,AE17Wei,REVISIT18Wei}, for the first aspect, we leverage a CNN for image classification, instead of exploiting off-the-shelf segmentation methods~(\emph{e.g.}, \cite{MCG16Pont,GRABCUT04Rother}). To this end, we propose a background-aware pooling~(BAP) method using an attention map, enabling discriminating foreground and background inside the bounding boxes. This allows to aggregate features within a foreground for image classification, while discarding those for a background, resulting in more accurate CAMs, rather than mainly highlighting the most discriminative parts~(\emph{e.g.},~faces in the person class) as in GAP~\cite{NIN13Lin,CAM16Zhou}. Specifically, we retrieve background regions inside the bounding boxes, based on our finding that background regions are perceptually consistent in part within an image. This provides attention maps for the background regions adaptively for individual images. We exploit the attention maps and CAMs, together with prototypical features, to generate pseudo ground-truth labels. For the second one, we introduce a noise-aware loss~(NAL) to train CNNs for semantic segmentation that makes the networks less susceptible to incorrect labels. Specifically, we exploit a confidence map, using the distances between CNN features for prediction and classifier weights for semantic segmentation, to compute a cross-entropy loss adaptively. Experimental results demonstrate that our approach to using BAP already outperforms the state of the art on the PASCAL VOC 2012 dataset~\cite{VOC10Ever}, and the NAL further boosts the performance. We also demonstrate the effectiveness of our approach by extending it to the task of instance segmentation on the MS-COCO dataset~\cite{COCO14Lin}. We summarize the contributions of our work as follows:
\begin{itemize}[leftmargin=*]
    \item[$\bullet$] We introduce a novel pooling method for WSSS, dubbed BAP, that uses bounding box annotations, allowing to generate high-quality pseudo ground-truth labels.
    \item[$\bullet$] We propose a NAL exploiting the distances between CNN features for prediction and classifier weights for semantic segmentation, lessening the influence of incorrect labels. 
    \item[$\bullet$] We set a new state of the art on the PASCAL VOC 2012 dataset for weakly- and semi-supervised semantic segmentation. We also provide an extensive experimental analysis with ablation studies.
\end{itemize}
Our code and models are available online:~\url{https://cvlab.yonsei.ac.kr/projects/BANA}.

\begin{figure*}
\centering
\includegraphics[width=\textwidth]{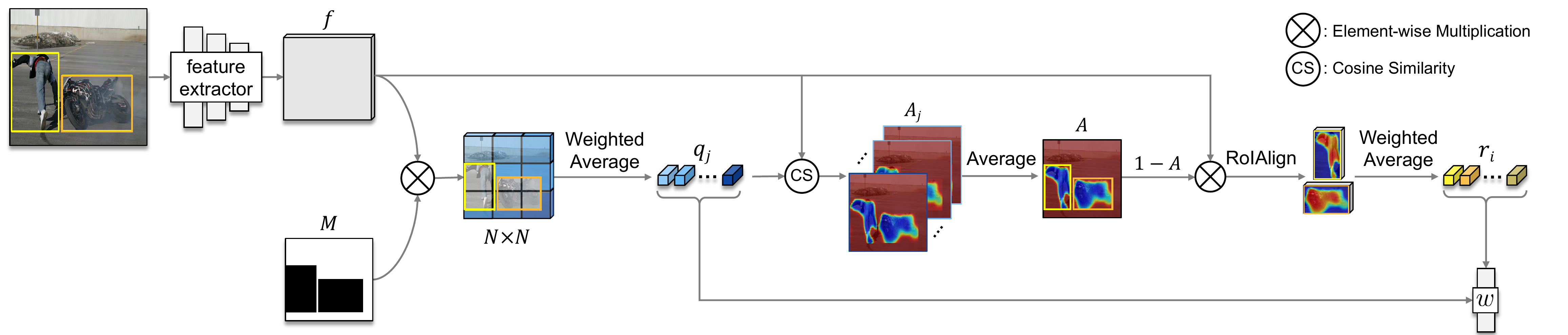}
\caption{\small{Overview of image classification using BAP. We first extract queries~$q_j$ using a feature map~$f$ and a binary mask~$M$ indicating a definite background. The queries~$q_j$ are then used to compute an attention map~$A$ describing the likelihood that each pixel belongs to a background. The attention map enables localizing entire foreground regions, leading to better foreground features~$r_i$. Finally, we apply a softmax classifier~$w$ to the foreground features~$r_i$ for each bounding box together with the queries~$q_j$. The entire network is trained with a cross-entropy loss. See text for details~(Sec.~\ref{sec:bap}). Best viewed in color.}}
\vspace{-0.3cm}
\label{fig:Fig2}
\end{figure*}

\section{Related work}
\noindent \textbf{WSSS using image-level labels.} Image-level labels have been used for WSSS as an alternative to dense pixel-level annotations. However, they indicate the presence or absence of objects of a particular class only, and do not provide any information for the location of objects, making the problem extremely challenging. The seminal work of~\cite{CAM16Zhou} proposes CAMs to estimate coarse localization maps for objects or actions by using a CNN trained with the image-level labels. Since then, several WSSS methods exploit CAMs to generate a pseudo ground truth, which is used to train CNNs for semantic segmentation in a supervised manner, by propagating them with DenseCRF~\cite{DSRG18Huang,SEC16Kol,BUILT16Sal,RDM19Zhang}, using a stochastic inference~\cite{FICKLE19Lee}, incorporating a self-supervised task~\cite{REBUTTAL120Chang}, or alternatively mining and erasing object-related regions~\cite{SeeNet18Hou,GAIN18Li,AE17Wei,REVISIT18Wei}. These approaches enlarge initial CAMs, typically activated on the most discriminative parts, progressively to cover entire objects, but this may highlight the regions irrelevant to objects~(\emph{e.g.},~a background). To address this problem, off-the-shelf saliency detectors~\cite{S4NET19Fan,DSOD17Hou,SOD13Jiang,ISOD17Li,SESOD17Xiao} and/or segment proposals~\cite{BING14Cheng,MCG16Pont} have been used, but they require calibrating saliency/objectness scores carefully. SSNet~\cite{JOINT19Zeng} proposes to learn WSSS and saliency detection jointly with a unified network architecture, but it requires ground-truth saliency annotations. Unlike these methods using image-level labels, we require no external models nor ground-truth saliency annotations. Recently, SeeNet~\cite{SeeNet18Hou} proposes to use background regions for WSSS. It defines a background explicitly using CAMs whose values are below a threshold, and prevents spreading CAMs into the background. Our work is similar to SeeNet in that both exploit the background regions. Differently, we define the background regions implicitly using a nonparametric retrieval technique, with an assumption that the background is perceptually consistent in part within an image. We then leverage them to design a new pooling method and to generate pseudo ground-truth labels. 

\noindent \textbf{WSSS using bounding box labels.} An alternative approach for WSSS is to exploit object bounding boxes as a supervisory signal. They are easy to annotate compared to pixel-level labels~(\emph{e.g.},~annotating boxes is about 15 times cheaper than labeling pixel-wise segments~\cite{COCO14Lin}), and provide a definite background with the extent of each object. In this context, recent methods~\cite{BOXSUP15Dai,SDI17Kho,BTS20Kul,WSSL15Pap,BOXDRIVEN19Song} close the performance gap between weakly-supervised and fully-supervised methods. For example, BoxSup~\cite{BOXSUP15Dai} employs MCG~\cite{MCG16Pont} to generate candidate segments and uses object bounding boxes to update the segments iteratively along with network parameters. WSSL~\cite{WSSL15Pap} adopts DenseCRF~\cite{DCRF11Kr} to generate pseudo segmentation labels using bounding boxes, and proposes an expectation-maximization algorithm to refine the labels. SDI~\cite{SDI17Kho} argues that generating correct pseudo labels is a crucial step for the performance of WSSS, and proposes to use GrabCut~\cite{GRABCUT04Rother} and MCG to estimate the labels. We also advocate the importance of high-quality pseudo labels for WSSS, but exploit a classification network, similar to the methods using image-level labels, instead of exploiting the off-the-shelf segmentation methods~\cite{MCG16Pont,GRABCUT04Rother} or additional datasets~\cite{BSDS10Arbel}.

\noindent \textbf{Learning from noisy labels.} Several methods have been proposed for correcting or filtering out noisy labels,~\emph{e.g.}, \cite{MENTORNET17Jiang,LNL13Nata,BOOTSTRAP14Reed,COMBINATORIAL19Seo,GCE18Zhang} to name a few. We refer to~\cite{SURVEY14Frenay} for a comprehensive review. In the context of WSSS, pseudo ground-truth labels generated by a weak form of supervisory signals are not accurate compared to manual annotations, making the task much more difficult. To alleviate the influence of noisy segmentation labels, SDI~\cite{SDI17Kho} discards regions whose labels obtained by MCG~\cite{MCG16Pont} and GrabCut~\cite{GRABCUT04Rother} are different, to train a network. Similarly, the work of~\cite{REBUTTAL220Fan} exploits multiple pseudo labels that might complement each other. This work, however, requires 12 different pseudo labels obtained from two classification networks. By contrast, we generate two pseudo labels with a negligible overhead. Recently, BCM~\cite{BOXDRIVEN19Song} introduces a filling rate constraint. It filters out incorrectly labeled pixels, based on the mean percentage of foreground pixels within bounding boxes of each object class. Box2Seg~\cite{BTS20Kul} also uses the filling rate constraint to regularize a class-specific attention map. The attention map is then used to modulate a cross-entropy loss to handle incorrect labels. These approaches~\cite{BTS20Kul,BOXDRIVEN19Song} outperform other methods, but require a pre-training stage to stabilize the training and use additional parameters to compute losses, which is in contrast to our NAL.

\section{Approach}
\begin{figure}[t]
\centering
\includegraphics[width=\linewidth]{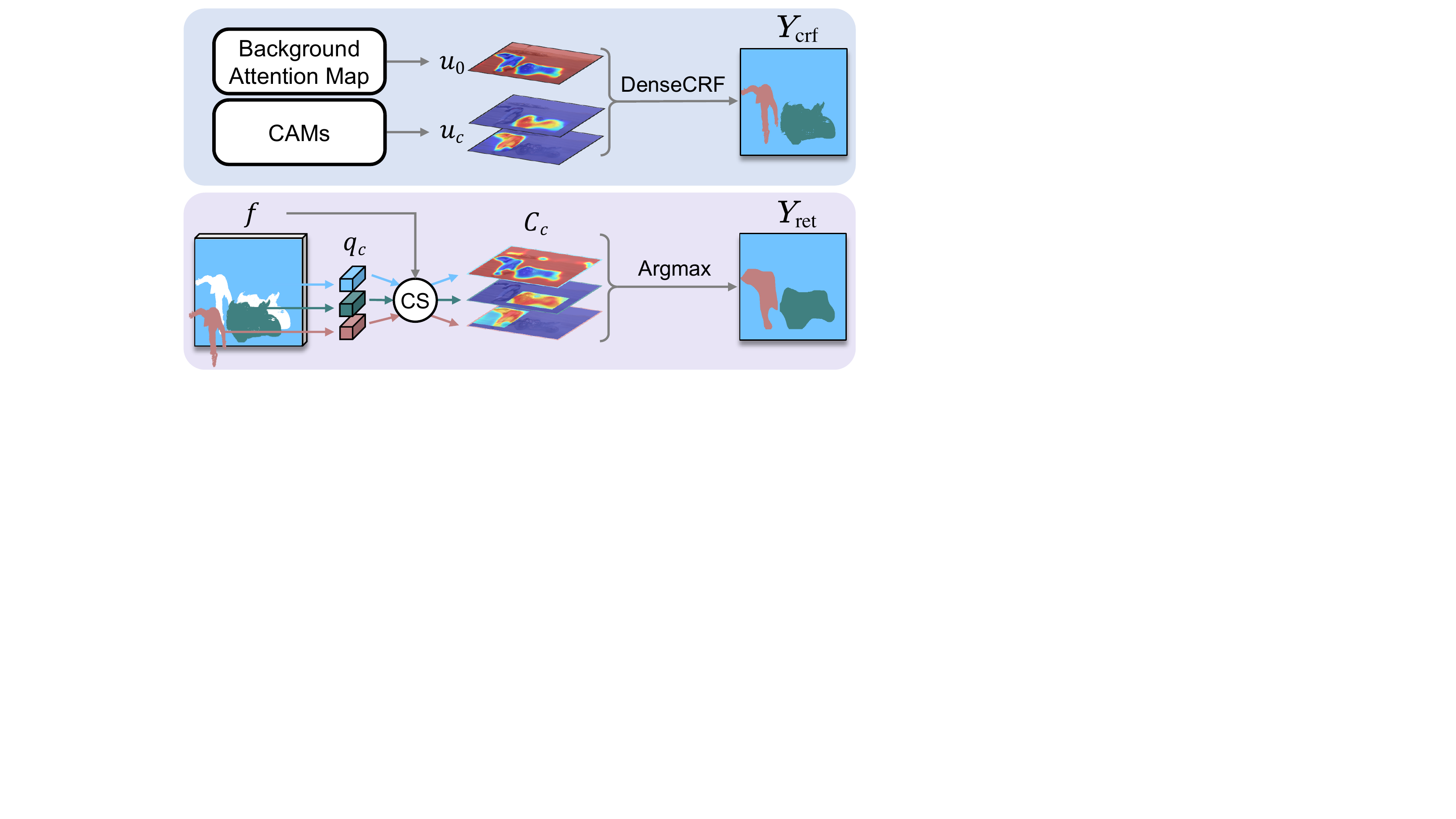}
\vspace{-0.7cm}
\caption{\small{Generating pseudo labels. We compute~$u_0$ and~$u_c$ using a background attention map and CAMs, respectively, which are used as a unary term for DenseCRF~\cite{DCRF11Kr} to obtain pseudo segmentation labels~$Y_\text{crf}$. We extract prototypical features~$q_c$ for each class using the labels~$Y_\text{crf}$, and use them as queries to retrieve high-level features from the feature map~$f$, from which we obtain additional pseudo labels~$Y_\text{ret}$. See text for detail~(Sec.~\ref{sec:generating}). Best viewed in color.}}
\vspace{-0.3cm}
\label{fig:Fig3}
\end{figure}

Our approach mainly consists of three stages: First, we train a CNN for image classification using object bounding boxes~(Fig.~\ref{fig:Fig2}). We use BAP leveraging a background prior, that is, background regions are perceptually consistent in part within an image, allowing to extract more accurate CAMs. To this end, we compute an attention map for a background adaptively for each image. Second, we generate pseudo segmentation labels using CAMs obtained from the classification network together with the background attention maps and prototypical features~(Fig.~\ref{fig:Fig3}). Finally, we train CNNs for semantic segmentation with the pseudo ground truth but possibly having noisy labels. We use a NAL to lessen the influence of the noisy labels. In the following, we describe a detailed description of each stage.
\subsection{Image classification using BAP} 
\label{sec:bap}
Our classification network consists of a feature extractor and a~($L+1$)-way softmax classifier~($L$ object classes and the background class). Given an input image, the feature extractor outputs a feature map~$f$. We denote by~$\mathcal{B} = \{B_1,B_2,\dots,B_K\}$ a set of object bounding boxes in the input image, where~$K$ is the number of bounding boxes, resized w.r.t. the size of the feature map~$f$ correspondingly using nearest-neighbor interpolation. We denote by~$M$ a mask indicating a definite background~(\emph{i.e.},~the regions outside the bounding boxes), where~$M(\mathbf{p})=1$ if the position~$\mathbf{p}$ does not belong to any bounding boxes, and~$M(\mathbf{p})=0$ otherwise.

\noindent \textbf{Background attention map.} Separating foreground and background regions inside object bounding boxes allows the classifier to focus more on learning foreground objects. As will be seen in our experiments, this results in better CAMs localizing entire objects. However, the object bounding boxes contain a mixture of foreground and background, and do not provide any information about object boundaries. To discriminate foreground and background regions for each bounding box, we pose this problem as a retrieval task. Specifically, we divide the feature map~$f$ into~$N\times N$ regular grids. We denote by~$G(j)$ each grid cell, where~$1 \leq j \leq N^2$. Note that we ignore invalid cells that do not overlap with the definite background regions at all,~(\emph{i.e.}, the cells inside object bounding boxes), suggesting that each input image could have different numbers of valid grid cells, at most~$N^2$. We then aggregate features for individual grid cells, and use them as queries for retrieval as follows:
\begin{equation} 
q_j = \frac{\displaystyle \sum_{\mathbf{p} \in G(j)} M(\mathbf{p}) f(\mathbf{p})}{\displaystyle \sum_{\mathbf{p} \in G(j)} M(\mathbf{p})}.
\end{equation}
That is, we obtain individual queries~$q_j$ by computing a weighted average of features~$f$ on corresponding grid cell~$G(j)$ using the binary mask~$M$. Given the queries, we retrieve the background regions inside the bounding boxes, and obtain an attention map~$A$ as follows:
\begin{equation}
	A(\mathbf{p}) = \frac{1}{J} \sum_{j} A_{j}(\mathbf{p}),
\end{equation}
where~$J$ is the number of valid grid cells and
\begin{equation}
A_{j}(\mathbf{p})= \left\{\begin{matrix}
\text{ReLU}\left(\frac{f(\mathbf{p})}{\|f(\mathbf{p})\|}\cdot
\frac{q_j}{\|q_j\|}\right)&,\mathbf{p} \in \mathcal{B} \\
1&,\mathbf{p} \notin \mathcal{B}
\end{matrix}\right..
\end{equation}
We denote by~$\|\cdot\|$ L2 normalization. This computes cosine similarity between features inside the bounding boxes~$\mathcal{B}$ and queries~$q_j$, and truncates the results into the range of~$[0,1]$ by the ReLU~\cite{ALEXNET12Kriz} function. Accordingly, the attention map~$A$ quantifies the likelihood that each pixel inside the bounding boxes belongs to a background. It is more likely to be a background, as the value of the attention map~$A$ approaches to one.

\noindent \textbf{BAP.} We use the attention map~$A$ to aggregate foreground features for each bounding box~$B_i$\footnote{We use the RoIAlign method~\cite{MASKRCNN17He} to extract features inside object bounding boxes.} as follows:
\begin{equation}
r_i = \frac{\displaystyle \sum_{\mathbf{p} \in B_i} (1-A(\mathbf{p})) f(\mathbf{p})}{\displaystyle \sum_{\mathbf{p} \in B_i} (1-A(\mathbf{p}))},
\end{equation}
which corresponds to weighted average pooling, where the weight is the probability of the point at~$\mathbf{p} \in B_i$ being a foreground. Note that it becomes GAP, when all regions inside the bounding box are considered as a foreground~(\emph{i.e.},~$A=0$).

\noindent \textbf{Loss.} 
We apply the~($L+1$)-way softmax classifier~$w$ to individual features for the foreground and background regions~(\emph{i.e.},~$r_i$ and~$q_j$, respectively) to train the classification network with a standard cross-entropy loss. This enables better distinguishing foreground objects from a background.

\subsection{Pseudo label generation}
\label{sec:generating}
We introduce two approaches, complementary to each other, to generate pseudo ground-truth labels. First, we leverage CAMs for each object bounding box obtained from the classification network using BAP. We exploit DenseCRF~\cite{DCRF11Kr} with a unary term for each object class~$c$ defined as:
\begin{equation}
\label{eq:u_c}
u_c(\mathbf{p}) = \left\{\begin{matrix}
\frac{\text{CAM}_c(\mathbf{p})}{\max_{\mathbf{p}}(\text{CAM}_c(\mathbf{p}))}&,
\mathbf{p} \in \mathcal{B}_c \\ 
0 & , \mathbf{p} \notin \mathcal{B}_c
\end{matrix}\right.,
\end{equation}
where we denote by~$\mathcal{B}_c$ a set of bounding boxes containing objects of the class~$c$ and
\begin{equation}
\label{eq:cam}
\text{CAM}_c(\mathbf{p}) = \text{ReLU}(f(\mathbf{p}) \cdot w_c).
\end{equation}
$w_c$ is the classifier weight for the object class~$c$. For the background class, we use the background attention map~$A$ as follows:
\begin{equation}
\label{eq:u_0}
u_0(\mathbf{p}) = A(\mathbf{p}).
\end{equation}
Note that we could also use the CAM for the background class directly, similar to other object classes, but it highlights the most frequently observed regions in the dataset during training, lessening the discriminative ability of CRF to separate foreground and background inside the bounding boxes. For a pairwise term, we follow other WSSS methods that use contrast-sensitive bilateral potentials using color values and positions as in~\cite{DCRF11Kr}. We concatenate the unary terms for each object class in Eq.~\eqref{eq:u_c} and the background in Eq.~\eqref{eq:u_0}, and input them to DenseCRF together with an input image to obtain segmentation labels~$Y_\text{crf}$. Second, while the first approach using DenseCRF delineates object boundaries, low-level features~(\emph{e.g.},~color and texture) in the pairwise term might result in incorrect segmentation labels. We thus leverage high-level features~$f$ obtained from the classification network to complement this. Specifically, we propose to use a retrieval technique similar to BAP. We extract a prototypical feature for each class as follows:
\begin{equation}
q_c = \frac{1}{|\mathcal{Q}_c|}{\displaystyle \sum_{\mathbf{p} \in \mathcal{Q}_c} f(\mathbf{p})},
\end{equation}
where~$\mathcal{Q}_c$ is a set of locations labeled as the class~$c$ in~$Y_\text{crf}$~(including the background class) and~$|\cdot|$ indicates the number of pixels. We use prototypical features as queries to retrieve similar ones from the feature map~$f$, and compute a correlation map for each class as follows:
\begin{equation}
	C_c(\mathbf{p}) = \frac{f(\mathbf{p})}{\|f(\mathbf{p})\|} \cdot \frac{q_c}{\|q_c\|}.
\end{equation}
We then obtain pseudo segmentation labels~$Y_\text{ret}$ by applying the argmax function over the correlation maps~$C_c$\footnote{We upsample~$u_c$,~$u_0$, and~$C_c$ by bilinear interpolation such that they have the same resolution as the input image.}.

\subsection{Semantic segmentation with noisy labels}
We train DeepLab~\cite{DEEPLAB14Chen,DEEPLAB18Chen} for semantic segmentation with the pseudo pixel-level labels,~$Y_\text{crf}$ and~$Y_\text{ret}$. We extract a feature map~$\phi$ from the penultimate layer, and pass it through a softmax classifier~$W$\footnote{Following~\cite{DYNAMIC18Gidaris,FEW18Qiao}, we use a cosine similarity based classifier that encourages the classifier weights to be more representative for the corresponding classes. More details are given in the supplementary material.}, resulting in a~$(L+1)$-dimensional probability map~$H$. To alleviate the influence of incorrect labels, we exploit the regions~$\mathcal{S}$, where both~$Y_\text{crf}$ and~$Y_\text{ret}$ give the same label, to compute the loss as follows: 
\begin{equation}
\label{eq:ce}
\mathcal{L}_\text{ce} = 
-\frac{1}{\sum_{c } |\mathcal{S}_c|}
\sum_{c} \sum_{\mathbf{p} \in \mathcal{S}_c} 
\log H_c(\mathbf{p}),
\end{equation}
where $H_c$ is a probability for the class~$c$, and $\mathcal{S}_c$ is a set of locations labeled as the class~$c$ in~$\mathcal{S}$. We also exploit other regions~${\sim}\mathcal{S}$, where~$Y_\text{crf}$ and~$Y_\text{ret}$ give different labels, rather than discarding them completely as in~\cite{SDI17Kho}. These regions are less reliable than~$\mathcal{S}$, but might contain correct labels. It is, however, hard to determine whether the label is correct or not. Motivated by the work of~\cite{SPHEREFACE17Liu,FEW18Qiao}, we assume that a classifier weight represents a center of each class in the feature space, suggesting that the weight can be thought of as a representative feature for the corresponding class. We thus distinguish the label noise by using the distances between CNN features and classifier weights. To implement this idea, we first compute a correlation map for each class as follows: 
\begin{equation}
\label{eq:correlation}
D_c(\mathbf{p})=1+\left(\frac{\phi(\mathbf{p})}{\|\phi(\mathbf{p})\|} \cdot
\frac{W_c}{\|W_c\|}\right),
\end{equation}
where we denote by~$W_c$ the classifier weight for the corresponding class~$c$. We use cosine similarity as a metric with adding one to force the correlation score to be positive. We then compute a confidence map as follows: 
\begin{equation}
\label{eq:confidence}
\sigma(\mathbf{p}) =
\left(\frac{D_{c^*}(\mathbf{p})}
{\max_c(D_c(\mathbf{p}))}\right)^{\gamma},
\end{equation}
where~$c^*$ is a label obtained by~$Y_\text{crf}$~(\emph{i.e.},~$c^* = Y_\text{crf}(\mathbf{p})$), and~$\gamma~(\geq 1)$ is a damping parameter. The confidence map provides the likelihood of each label being correct. The rationale for this is that the correlation values of $D_{c^*}(\mathbf{p})$ and $\max_c(D_c(\mathbf{p}))$ will be similar, when the label~$c^*$ is confident, and vice versa. Note that we can adjust the confidence values with the damping parameter~$\gamma$. When~$\gamma$ approaches to infinity, the values become binary, considering the most confident labels only and acting as a hard constraint. That is,~$\sigma(\mathbf{p})=1$ only when~$D_{c^*}(\mathbf{p}) = \max_c(D_c(\mathbf{p}))$, and~$\sigma(\mathbf{p}) \approx 0$ otherwise. We exploit the confidence map as a weighting factor to compute the cross-entropy loss as follows:
\begin{equation}
\label{eq:wce}
\mathcal{L}_\text{wce} = -\frac{1}
{\sum_{c} \sum_{\mathbf{p} \in \mathcal{{\sim}S}_c} \sigma(\mathbf{p})}
\sum_{c} \sum_{\mathbf{p} \in \mathcal{{\sim}S}_c}
\sigma(\mathbf{p})\,\log H_c(\mathbf{p}),
\end{equation}
where we denote by~${\sim}\mathcal{S}_c$ a set of location labeled as the class~$c$ in~${\sim}\mathcal{S}$. Accordingly, the overall NAL is defined with a balance parameter~$\lambda$ as follows:
\begin{equation}
\mathcal{L} = \mathcal{L}_\text{ce} + \lambda \, \mathcal{L}_\text{wce}.
\end{equation}

\section{Experiments}
In this section, we describe implementation details, and present a detailed analysis of our method with ablation studies. We then compare our model with state-of-the-art WSSS methods. We obtain experimental results using~$\texttt{PyTorch}$~\cite{PYTORCH17Pas} with a NVIDIA Titan RTX GPU. More results including qualitative comparisons can be found in the supplementary material. 

\subsection{Implementation details}
\noindent \textbf{Dataset and evaluation.}
We use the PASCAL VOC 2012 dataset~\cite{VOC10Ever} consisting of~$1,464$/$1,449$/$1,456$ samples of 21 classes~(including the background class) for~$train$,~$val$, and~$test$, respectively. Following the common practice in~\cite{BOXSUP15Dai,SEC16Kol,FICKLE19Lee,BOXDRIVEN19Song}, we use augmented~$10,582$ training samples provided by~\cite{SBD11Hari} to train our models. We use the mean intersection-over-union~(mIoU) metric to measure the precision of pseudo segmentation labels and segmentation results. We obtain results for the~$test$ set on the official PASCAL VOC evaluation server. For instance segmentation, we use MS-COCO~\cite{COCO14Lin} for 115K/5K/20K samples of~$train$,~$val$, and~$test$, respectively, containing 81 classes including the background class. We use the average precision~(AP) metrics to evaluate pseudo labels and segmentation results.

\noindent \textbf{Classification network.}
We adopt the classification network in AffinityNet~\cite{AFFINITY18Ahn}, a slight modification of VGG-16~\cite{VGG15Simo} pre-trained for ImageNet classification~\cite{IMAGENET09Deng}, to extract the feature map~$f$. Initially, classifier weights are sampled randomly from a Gaussian distribution with zero mean and standard deviation of 1e-2. We train the classification network for 15 epochs with a batch size of 20 using the SGD optimizer with momentum of~$0.9$ and weight decay of 5e-4. Learning rates are initially set to 1e-4 and 1e-3 for the pre-trained layers and the classifier, respectively, and they are divided by 10 after 10 epochs. We augment the training set with horizontal flipping, random cropping~($321\times321$), random scaling, and color jittering.

\noindent \textbf{Segmentation network.}
For fair comparison, we exploit two models for semantic segmentation: DeepLab-V1~(LargeFOV)~\cite{DEEPLAB14Chen,DEEPLAB18Chen} with VGG-16~\cite{VGG15Simo} and DeepLab-V2~(ASPP)~\cite{DEEPLAB18Chen} with ResNet-101~\cite{RESNET16He}. We train DeepLab-V1~(V2) for 45~(20) epochs with a batch size of 20~(10), and adjust the learning rate using the poly schedule~\cite{DEEPLAB18Chen}.

\noindent \textbf{Hyperparameter settings.}
Following the experimental protocol in~\cite{BOXSUP15Dai,WSSL15Pap}, parameters for DenseCRF~\cite{DCRF11Kr} are chosen by cross-validation on the held-out set of 100 validation images fully-annotated. We set the grid size $N$ to 4 for training a classification network using BAP in order to see diverse background queries. We empirically find that using confident background regions in Eq.~\eqref{eq:u_0}, that is, thresholding $A$ with $0.99$, and setting $N$ to 1 provide better results when generating pseudo labels. We use a grid search to set the threshold value for $A$ and the grid size $N$ on the same held-out set for DenseCRF. For DeepLab, we use default settings in~\cite{DEEPLAB14Chen,DEEPLAB18Chen}. Other parameters are fixed to all experiments ($\gamma=7,~\lambda=0.1$). In the supplementary material, we provide quantitative comparisons and more analysis on these parameters. 

\subsection{Analysis}
\begin{table}[t]
\centering
\footnotesize
\caption{\small{Comparison of pseudo labels on the PASCAL VOC 2012~\cite{VOC10Ever} $train$ and $val$ sets in terms of mIoU. Numbers in bold indicate the best performance. We report the supervision types with the number of annotations. For MCG~\cite{MCG16Pont}, we manually choose the segment proposal that gives the highest IoU score with each bounding box. $^*$: pseudo labels contain unreliable regions.}}
\vspace{-0.3cm}
\label{table:pseudo-voc}
\begin{tabular}{c c c}
\specialrule{.1em}{.05em}{.05em}
\multicolumn{1}{l}{Method} & $train$ & $val$
\\ 
\hline
\hline
\multicolumn{2}{l}{\textit{Supervision}: Image-level labels~(10K)}
\\
\multicolumn{1}{l}{AffinityNet$_{\text{CVPR'18}}$~\cite{AFFINITY18Ahn}} 
& \multicolumn{1}{c}{59.7} & \multicolumn{1}{c}{-}
\\
\hline
\multicolumn{1}{l}{\textit{Supervision}: Boxes~(10K)}
\\
\multicolumn{1}{l}{Box} & \multicolumn{1}{c}{65.4} & \multicolumn{1}{c}{62.2}
\\
\multicolumn{1}{l}{GrabCut$_{\text{TOG'04}}$~\cite{GRABCUT04Rother}}
& \multicolumn{1}{c}{65.7} & \multicolumn{1}{c}{66.1}
\\
\multicolumn{1}{l}{MCG$_{\text{PAMI'16}}$~\cite{MCG16Pont}}
& \multicolumn{1}{c}{66.2} & \multicolumn{1}{c}{66.9}
\\
\multicolumn{1}{l}{WSSL$_{\text{ICCV'15}}$~\cite{WSSL15Pap}}
& \multicolumn{1}{c}{69.7} & \multicolumn{1}{c}{71.1}
\\
\multicolumn{1}{l}{SDI$_{\text{CVPR'17}}$~\cite{SDI17Kho}}
& \multicolumn{1}{c}{79.7$^*$} & \multicolumn{1}{c}{60.5$^*$}
\\
\multicolumn{1}{l}{Ours}
\\
\multicolumn{1}{l}{~~~~GAP}
& \multicolumn{1}{c}{75.5} & \multicolumn{1}{c}{76.1}
\\
\multicolumn{1}{l}{~~~~BAP:~$Y_{\text{crf}}$ w/o~$u_0$}
& \multicolumn{1}{c}{77.0} & \multicolumn{1}{c}{77.8}
\\
\multicolumn{1}{l}{~~~~BAP:~$Y_{\text{crf}}$}
& \multicolumn{1}{c}{\textbf{78.7}} & \multicolumn{1}{c}{\textbf{79.2}}
\\
\multicolumn{1}{l}{~~~~BAP:~$Y_{\text{ret}}$}
& \multicolumn{1}{c}{70.8} & \multicolumn{1}{c}{69.9}
\\
\multicolumn{1}{l}{~~~~BAP:~$Y_{\text{crf}}$ \&~$Y_{\text{ret}}$}
& \multicolumn{1}{c}{85.3$^*$} & \multicolumn{1}{c}{68.2$^*$}
\\
\specialrule{.1em}{.05em}{.05em}	
\end{tabular}
\vspace{-0.45cm}
\end{table}

\begin{table}[t]
\centering
\footnotesize
\caption{\small{Comparison of pseudo labels on the MS-COCO~\cite{COCO14Lin} $train$ set. Note that the results for `VOC-to-COCO' do not use any samples in the MS-COCO~$train$ set during training.}}
\vspace{-0.3cm}
\label{table:pseudo-coco}
\newcolumntype{C}[1]{>{\centering\arraybackslash}p{#1}}
\begin{tabular}{c C{0.45cm} C{0.45cm} C{0.45cm} C{0.45cm} C{0.45cm} C{0.45cm}}
\specialrule{.1em}{.05em}{.05em}
\multicolumn{1}{l}{Method} & $\text{AP}$ & $\text{AP}_{50}$ & $\text{AP}_{75}$ & $\text{AP}_S$ & $\text{AP}_M$ & $\text{AP}_L$
\\ 
\hline
\hline
\multicolumn{1}{l}{VOC-to-COCO}
\\
\multicolumn{1}{l}{~~BAP:~$Y_{\text{crf}}$} & 11.7 & 28.7 & 8.0 & 3.0 & 15.0 & 27.1
\\
\multicolumn{1}{l}{~~BAP:~$Y_{\text{ret}}$} & 9.0 & 30.1 & 2.8 & 4.4 & 10.2 & 16.2
\\
\hline
\multicolumn{1}{l}{COCO-to-COCO}
\\
\multicolumn{1}{l}{~~BAP:~$Y_{\text{crf}}$} & 17.2 & 40.5 & 12.5 & 5.9 & 20.4 & 32.2
\\
\multicolumn{1}{l}{~~BAP:~$Y_{\text{ret}}$} & 17.2 & 49.7 & 7.6 & 12.0 & 17.1 & 22.5
\\
\specialrule{.1em}{.05em}{.05em}
\end{tabular}
\vspace{-0.45cm}
\end{table}
\noindent \textbf{Accuracy of pseudo labels.} We compare in Table~\ref{table:pseudo-voc} mIoU scores of pseudo segmentation labels on the PASCAL VOC 2012~\cite{VOC10Ever}~$train$ and~$val$ sets. Note that our pseudo label generator can segment foreground objects given bounding boxes, even for unseen images during training. To the baseline, we consider the bounding box itself as a foreground object~(`Box'), which outperforms AffinityNet~\cite{AFFINITY18Ahn} using image-level labels. This suggests that the bounding box is a strong indicator for segmenting objects. WSSL~\cite{WSSL15Pap} using DenseCRF~\cite{DCRF11Kr} outperforms the baseline and hand-crafted methods~\cite{MCG16Pont,GRABCUT04Rother} by a considerable margin. To validate our approach to using a classification network with bounding boxes, we train a CNN with GAP for image classification, and generate pseudo labels using CAMs and DenseCRF~(`GAP'). We can clearly see that this approach already outperforms WSSL significantly, demonstrating the effectiveness of our approach. GAP, however, does not discriminate foreground and background regions inside bounding boxes. BAP overcomes this problem, and provides better CAMs, resulting in more accurate pseudo labels~(`BAP:~$Y_\text{crf}$'). Note that BAP does not introduce any additional parameters, similar to GAP. For comparison, we generate pseudo labels by replacing the background attention map $u_0$ in~Eq.~\eqref{eq:u_0} with the CAM for the background class~(`BAP:~$Y_\text{crf}$~w/o~$u_0$'), the performance of which is lower than `BAP:~$Y_\text{crf}$'. A plausible explanation is that the CAM for the background class highlights the most frequently observed regions only. In contrast to this, the attention map~$u_0$, a by-product of our BAP, marks background regions adaptively for individual images. We also report the mIoU scores for the pseudo labels obtained by a retrieval technique~(`BAP:~$Y_\text{ret}$'). Although this provides worse results than `BAP:~$Y_\text{crf}$', due to the use of the low-resolution feature map~$f$, as will be shown later, they are complementary to each other. Note that both pseudo labels of SDI~\cite{SDI17Kho} and ours~(`BAP:~$Y_\text{crf}$ \&~$Y_\text{ret}$') contain unreliable regions as shown in Fig.~\ref{fig:teaser}, making it hard to compare them with other methods. `BAP:~$Y_\text{crf}$ \&~$Y_\text{ret}$' show better results than SDI, but this should not be considered as fair comparison since performance would be different depending on the quantity of unreliable regions.

Our pseudo label generator is generic in that the adaptive attention map~$u_0$ allows to segment foreground objects for unseen classes during training. For example, even if we do not have a classifier weight for a novel class~(see~Eq.~\eqref{eq:cam}), we can exploit~$1-u_0$ as a class-agnostic foreground attention map. To validate this, we perform a cross-dataset evaluation in Table~\ref{table:pseudo-coco}, where we generate pseudo labels on the MS-COCO~\cite{COCO14Lin}~$train$ set by using the generator trained on PASCAL VOC 2012~\cite{VOC10Ever}~(`VOC-to-COCO'). For comparison, we train and evaluate our generator on the same dataset~(`COCO-to-COCO'). From this table, we observe three things:~(1) VOC-to-COCO gives reasonable pseudo labels even though it does not use any training samples of COCO during training. This is because our model computes the attention map adaptively, allowing to handle unseen object classes;~(2) Two pseudo labels,~$Y_{\text{crf}}$ and~$Y_{\text{ret}}$, are complementary to each other. For example,~$Y_{\text{ret}}$ gives better results for small objects, while~$Y_{\text{crf}}$ performs better for large objects;~(3) COCO-to-COCO provides better results, demonstrating the flexibility of our approach.

\begin{figure}[t]
\captionsetup[subfigure]{labelformat=empty}
\centering
\tiny
\hspace{-0.1cm}
\begin{minipage}{.54\linewidth}
\begin{subfigure}{.3\textwidth}
  \centering
  \frame{\includegraphics[width=\textwidth]{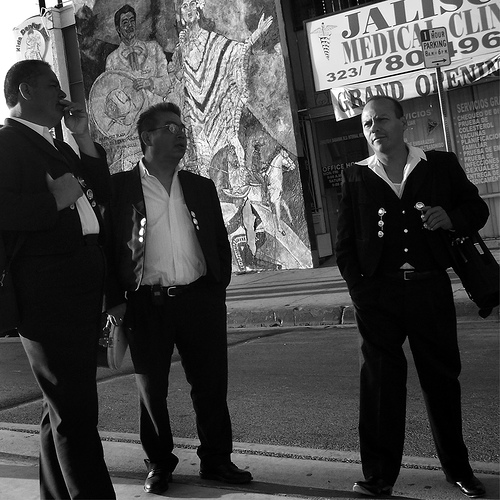}}
\end{subfigure}
\begin{subfigure}{.3\textwidth}
  \centering
  \frame{\includegraphics[width=\textwidth]{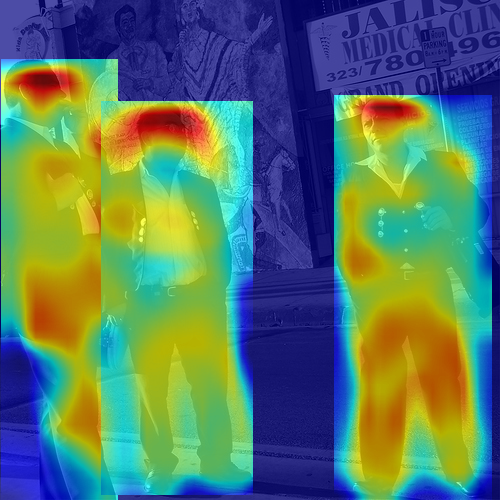}}
\end{subfigure}
\begin{subfigure}{.3\textwidth}
  \centering
  \frame{\includegraphics[width=\textwidth]{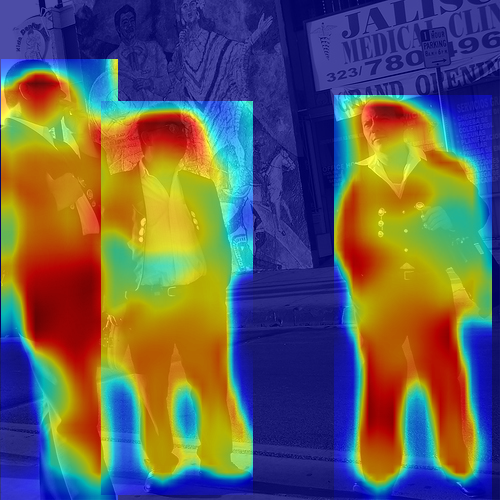}}
\end{subfigure}

\begin{subfigure}{.3\textwidth}
  \centering
  \frame{\includegraphics[width=\textwidth]{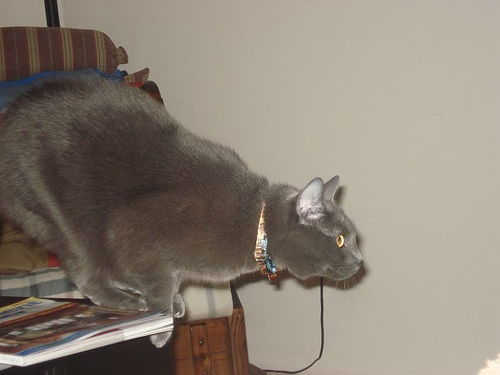}}
\end{subfigure}
\begin{subfigure}{.3\textwidth}
  \centering
  \frame{\includegraphics[width=\textwidth]{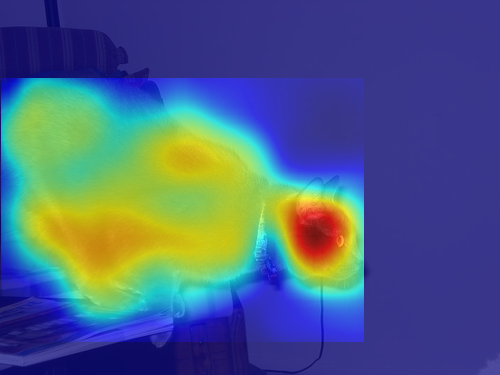}}
\end{subfigure}
\begin{subfigure}{.3\textwidth}
  \centering
  \frame{\includegraphics[width=\textwidth]{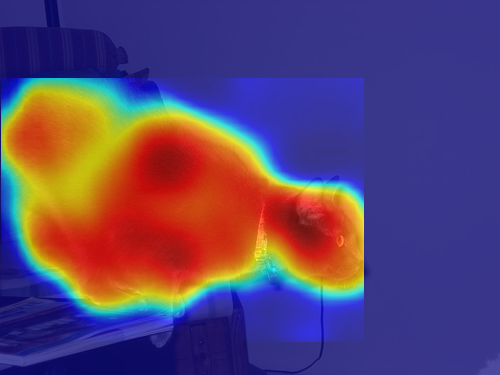}}
\end{subfigure}

\begin{subfigure}{.3\textwidth}
  \centering
  \frame{\includegraphics[width=\textwidth]{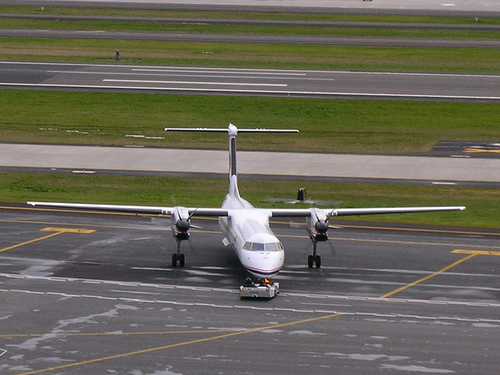}}\vspace{-0.2cm}
  \caption{\footnotesize Input.}
\end{subfigure}
\begin{subfigure}{.3\textwidth}
  \centering
  \frame{\includegraphics[width=\textwidth]{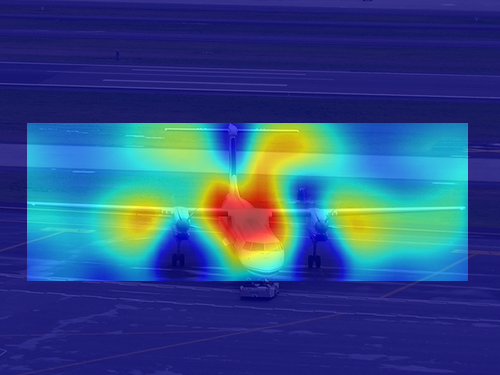}}\vspace{-0.2cm}
  \caption{\footnotesize GAP.}
\end{subfigure}
\begin{subfigure}{.3\textwidth}
  \centering
  \frame{\includegraphics[width=\textwidth]{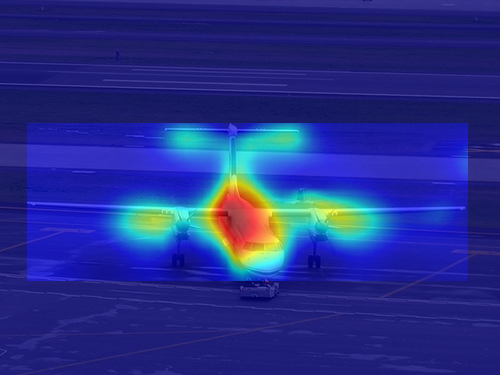}}\vspace{-0.2cm}
  \caption{\footnotesize BAP.}
\end{subfigure}
\end{minipage}
\hspace{-0.25cm}
\begin{minipage}{.46\linewidth}
  \centering
  \includegraphics[width=\textwidth]{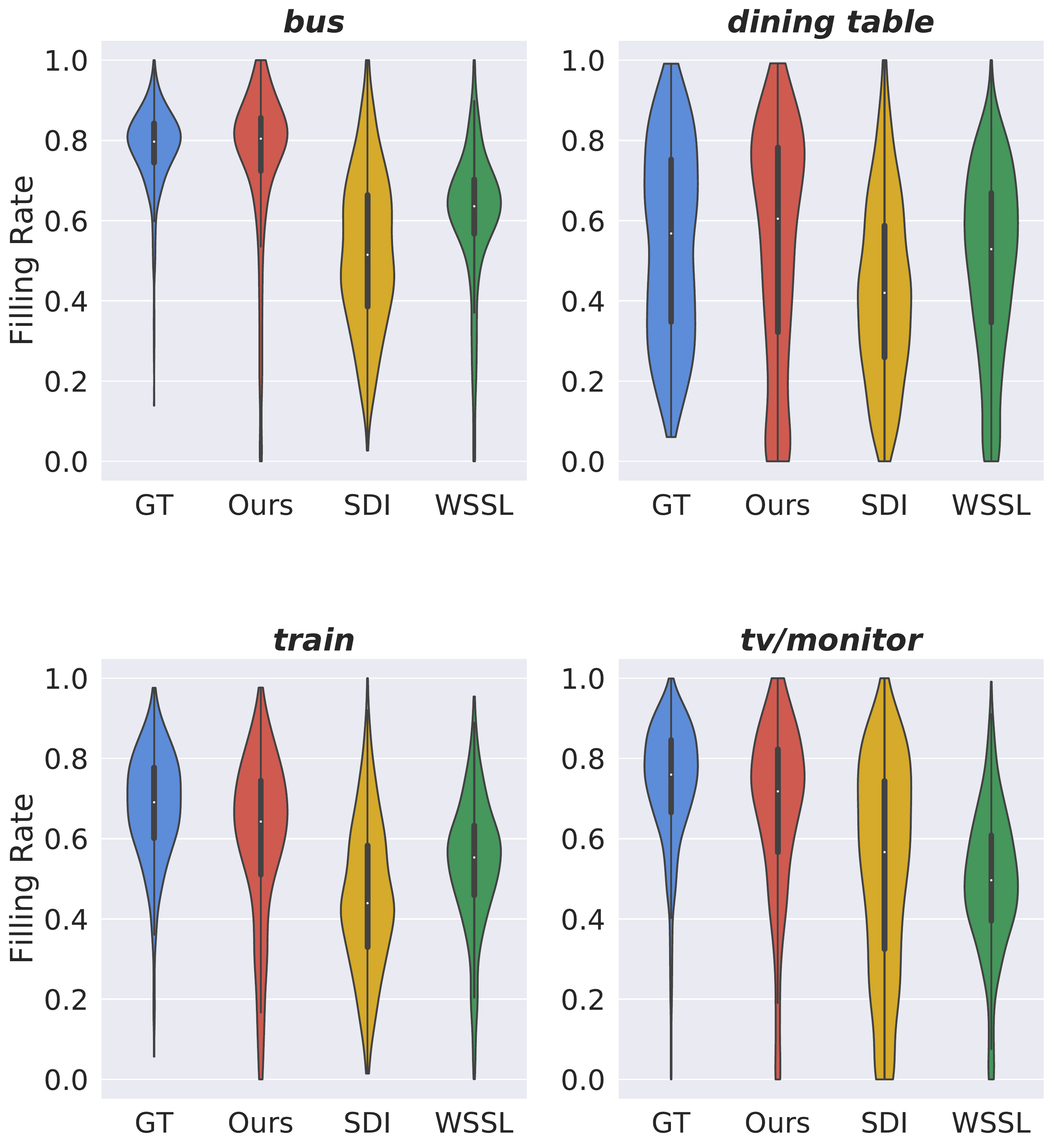}
\end{minipage}

\begin{minipage}{.5\linewidth}
  \centering
  \small{(a) CAMs.}
\end{minipage}
\begin{minipage}{.49\linewidth}
  \centering
  \small{~~~(b) Violin plots of filling rate.}
\end{minipage}
\vspace{-.2cm}
\caption{\small{(a) Visual comparison of CAMs using GAP and BAP. Our model using BAP provides better CAMs, highlighting the entire objects while suppressing irrelevant regions. (b) Comparison of filling rate distributions on the PASCAL VOC 2012~\cite{VOC10Ever}~$train$ set. Best viewed in color.}}
\label{fig:Fig4}
\vspace{-.45cm}
\end{figure}

\noindent \textbf{Comparison of CAMs.} To verify that our BAP improves the quality of CAMs, we provide visual examples of CAMs in Fig.~\ref{fig:Fig4}(a). We can clearly see that our BAP localizes the entire extent of objects~(\emph{e.g.}, man's legs in the first example) and does not highlight the irrelevant regions~(\emph{e.g.}, airstrip in the third example). Quantitative comparisons of BAP and GAP in terms of the precision of CAMs and classification performance can be found in the supplementary material. 

\noindent \textbf{Comparison of filling rates.} Recent methods~\cite{BTS20Kul,BOXDRIVEN19Song} show that a filling rate, the percentage of foreground pixels inside the bounding box, can be a good indicator to select the most confidence regions for back propagation. They use GrabCut~\cite{GRABCUT04Rother} to generate pseudo labels, and adopt the labels from WSSL~\cite{WSSL15Pap}, respectively. The per-class filling rates computed with these methods, however, vary significantly, and they are far from the ground truth. We show in Fig.~\ref{fig:Fig4}(b) examples of violin plot distributions for filling rates. We compare the filling rates estimated by SDI~\cite{SDI17Kho}, WSSL, ours, and the ground truth. We can see that the filling rates generated by our pseudo labels are more closer to the ground truth than other methods. We expect that our pseudo labels could improve the performance of other WSSS methods using the filling rate\footnote{Since these methods~\cite{BTS20Kul,BOXDRIVEN19Song} do not provide the source code at the time of submission, we could not perform this experiment.}.
 
\begin{table}[t]
\centering
\footnotesize
\caption{\small{Comparison of mIoU scores using different losses for the regions,~${\sim}\mathcal{S}$, where~$Y_\text{crf}$ and~$Y_\text{ret}$ give different labels, on the PASCAL VOC 2012~\cite{VOC10Ever}~$val$ set. We provide both mIoU scores before/after applying DenseCRF~\cite{DCRF11Kr}.}}
\vspace{-0.3cm}
\label{table:nal}
\begin{tabular}{c c}
\specialrule{.1em}{.05em}{.05em}
\multicolumn{1}{l}{Method} & $val$
\\
\hline
\hline
\multicolumn{1}{l}{Baseline} & \multicolumn{1}{c}{61.8 / 67.5}
\\
\multicolumn{1}{l}{~w/ Entropy Regularization~\cite{EM05Grand}} & \multicolumn{1}{c}{61.4 / 67.3}
\\
\multicolumn{1}{l}{~w/ Bootstrapping~\cite{BOOTSTRAP14Reed}} & \multicolumn{1}{c}{61.9 / 67.6}
\\
\multicolumn{1}{l}{~w/~$\mathcal{L}_\text{wce}$~(Eq.~$\eqref{eq:wce}$)} & \multicolumn{1}{c}{\textbf{62.4} / \textbf{68.1}}
\\
\specialrule{.1em}{.05em}{.05em}
\end{tabular}
\vspace{-0.3cm}
\end{table}

\begin{figure}[t]
\captionsetup[subfigure]{labelformat=empty}
\centering
\tiny
\begin{subfigure}{.117\textwidth}
  \centering
  \frame{\includegraphics[width=\textwidth,height=.8\textwidth]{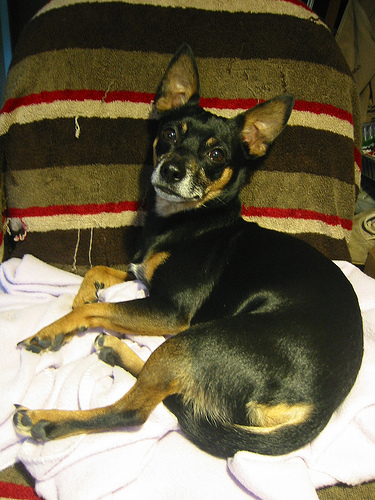}}
  \caption{Input image.}
\end{subfigure}
\begin{subfigure}{.117\textwidth}
  \centering
  \frame{\includegraphics[width=\textwidth,height=.8\textwidth]{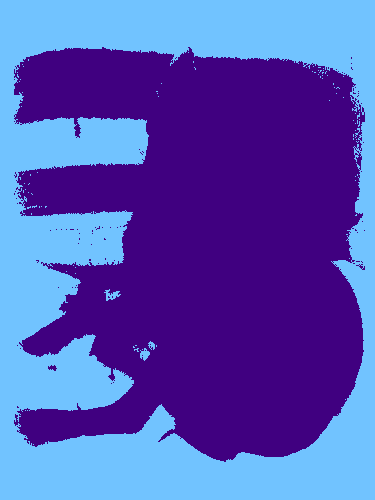}}
  \caption{Ours.}
\end{subfigure}
\begin{subfigure}{.117\textwidth}
  \centering
  \frame{\includegraphics[width=\textwidth,height=.8\textwidth]{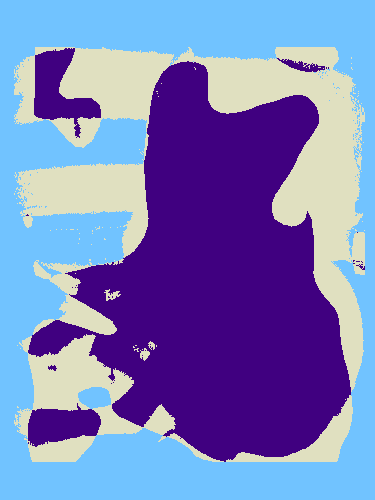}}
  \caption{Ours$^*$.}
\end{subfigure}
\begin{subfigure}{.117\textwidth}
  \centering
  \frame{\includegraphics[width=\textwidth,height=.8\textwidth]{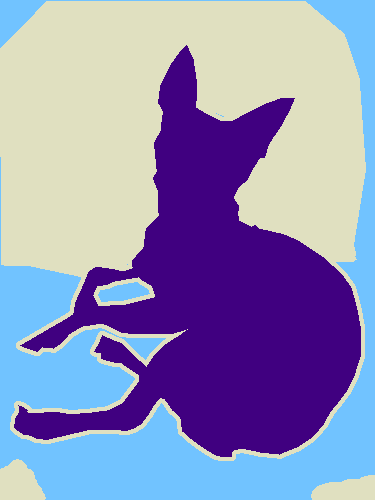}}
  \caption{Ground truth.}
\end{subfigure}

\begin{subfigure}{.09\textwidth}
  \centering
  \frame{\includegraphics[width=\textwidth,height=.8\textwidth]{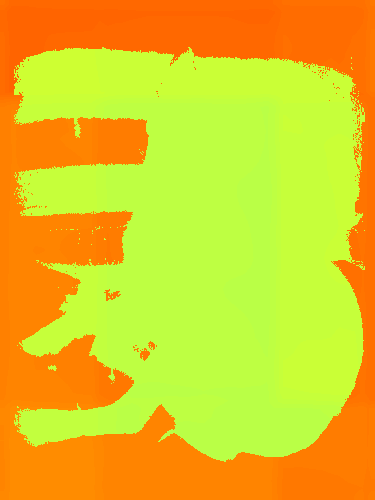}}
\end{subfigure}
\begin{subfigure}{.09\textwidth}
  \centering
  \frame{\includegraphics[width=\textwidth,height=.8\textwidth]{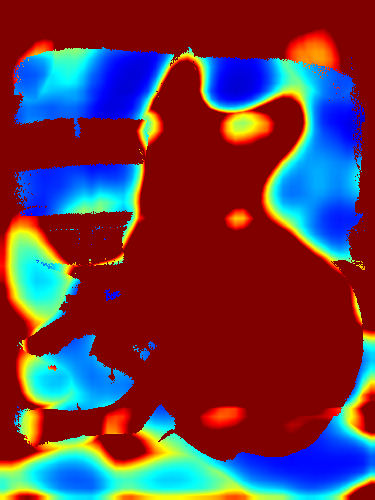}}
\end{subfigure}
\begin{subfigure}{.09\textwidth}
  \centering
  \frame{\includegraphics[width=\textwidth,height=.8\textwidth]{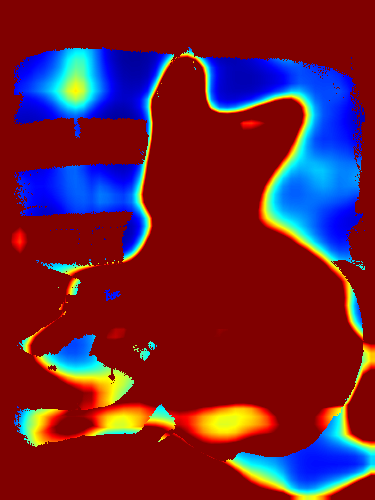}}
\end{subfigure}
\begin{subfigure}{.09\textwidth}
  \centering
  \frame{\includegraphics[width=\textwidth,height=.8\textwidth]{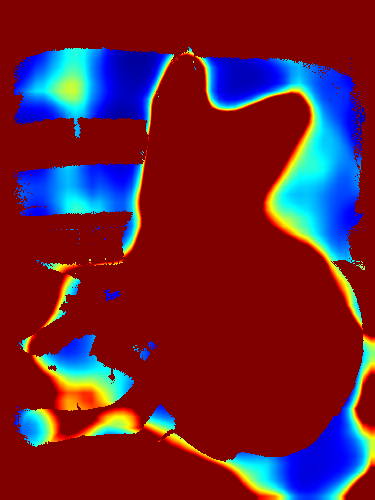}}
\end{subfigure}
\begin{subfigure}{.09\textwidth}
  \centering
  \frame{\includegraphics[width=\textwidth,height=.8\textwidth]{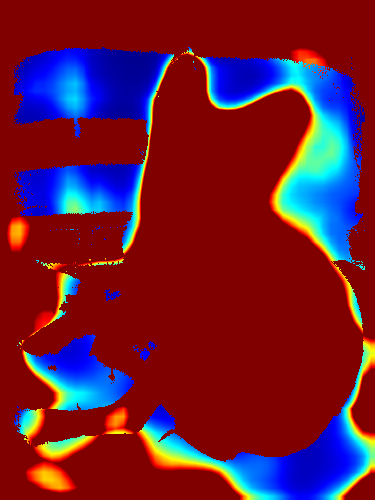}}
\end{subfigure}
\begin{subfigure}{.015\textwidth}
  \centering
  \includegraphics[width=\textwidth,height=5.4\textwidth]{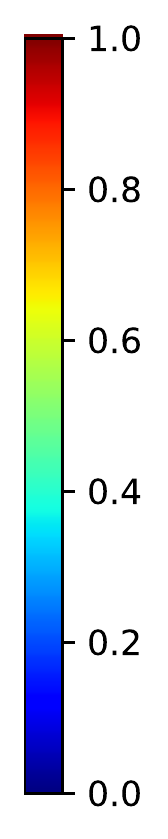}
\end{subfigure}
\begin{minipage}{\linewidth}
\centering
\small{Confidence maps~$\sigma$.}
\end{minipage}
\vspace{-0.25cm}
\caption{\small{Visual comparison of our pseudo labels $Y_\text{crf}$, the same ones but with an indication of unreliable regions using $Y_\text{ret}$, and the ground truth for the input image (top). Visualization of confidence maps at (from left to right) 0, 11, 22, 33 and 45 epochs (bottom). Best viewed in color.}}
\label{fig:Fig5}
\vspace{-0.49cm}
\end{figure}

\noindent \textbf{NAL.} We show in Table~\ref{table:nal} mIoU scores of DeepLab-V1~\cite{DEEPLAB14Chen} using different losses for regions~${\sim}\mathcal{S}$, where~$Y_\text{crf}$ and~$Y_\text{ret}$ give different labels. To the baseline, we ignore~${\sim}\mathcal{S}$ completely as in~\cite{SDI17Kho}. We can see that a bootstrapping technique~\cite{BOOTSTRAP14Reed} boosts the mIoU performance slightly, while an entropy regularization method~\cite{EM05Grand} does not. Our NAL penalizes incorrect labels adaptively, achieving the best result. This suggests that some pseudo labels in the regions~${\sim}\mathcal{S}$ are correct and exploiting them can boost the performance. We compare in Fig.~\ref{fig:Fig5} our pseudo labels with the ground truth, and visualize an evolution of the confidence map in Eq.~\eqref{eq:confidence} during training. As the input image contains two different objects having similar color,~\emph{i.e.}, dark stripes and a dog, our pseudo labels~$Y_{\text{crf}}$ are incorrect for the stripes, which could be, however, marked with~$Y_{\text{ret}}$. We can also see that our NAL assigns low confidence scores to the stripes, while maintaining high scores for the dog's legs.

\subsection{Segmentation results}
\noindent \textbf{PASCAL VOC.} We compare in Table~\ref{table:quantitative-vgg} the mIoU performance of our approach and state-of-the-art methods using DeepLab-V1~\cite{DEEPLAB14Chen,DEEPLAB18Chen}. We can see that our model trained with~$Y_{\text{crf}}$ already outperforms the state of the art by a significant margin without altering the training scheme~\cite{DEEPLAB14Chen,DEEPLAB18Chen}. This demonstrates that our approach to using a classification network with bounding boxes could be a promising way to generate a pseudo ground truth. We can also see that exploiting both labels,~$Y_{\text{crf}}$ and~$Y_{\text{ret}}$, together with our NAL further boosts the performance. We also report results for semi-supervised semantic segmentation. Following the experimental protocol in~\cite{BOXSUP15Dai,SDI17Kho,WSSL15Pap,BOXDRIVEN19Song}, we use ground-truth segmentation labels of~$1,464$ images in the original~$train$ set (13.8\% of the total images). Our method again achieves the best performance on the~$val$ and~$test$ sets. 

We report in Table~\ref{table:quantitative-res} segmentation results using DeepLab-V2~\cite{DEEPLAB18Chen}. It shows that we can achieve higher mIoU scores with a deeper CNN. The behavior of mIoU scores is almost the same as the one for DeepLab-V1 in Table~\ref{table:quantitative-vgg}. Our model trained with~$Y_{\text{ret}}$ even outperforms BCM~\cite{BOXDRIVEN19Song} in this case. This suggests that our pseudo labels are more likely to be exploited by deeper CNNs, further demonstrating the importance of high-quality pseudo labels for WSSS. Although Box2Seg~\cite{BTS20Kul} gives the best performance, it adopts UPerNet~\cite{UPer18Xiao} which requires a feature pyramid network~\cite{FPN17Lin}~(FPN), a pyramid pooling module~\cite{PPM17Zhao}~(PPM), and additional decoders. Note that UPerNet yields performance similar to PSPNet~\cite{PPM17Zhao} that outperforms DeepLab-V2 significantly.

\noindent \textbf{MS-COCO.} 
We show in Table~\ref{table:coco} a quantitative comparison of our models and other methods for instance segmentation. To this end, we train Mask-RCNN~\cite{MASKRCNN17He} with our pseudo labels. Since Mask-RCNN uses a binary cross-entropy loss for each object class, we could not obtain the correlation map in~Eq.~\eqref{eq:correlation}. We thus compute the loss for the regions~$\mathcal{S}$ only. For comparison, we report results of Mask-RCNN trained with ground-truth segmentation labels. From this table, we can see that Mask-RCNN trained with VOC-to-COCO outperforms AISI~\cite{ASSOCIATE18Fan}, demonstrating that the generalization ability of our method. Note that AISI generates a pseudo ground truth using image-level labels, but requires the instance-level saliency detector~\cite{S4NET19Fan} trained with ground-truth saliency annotations. Our model trained on COCO-to-COCO outperforms the other one using VOC-to-COCO, demonstrating again the importance of high-quality pseudo labels~(see the results in Table~\ref{table:pseudo-coco}).

\begin{table}[t]
\centering
\footnotesize
\caption{\small{Quantitative comparison with state-of-the-art methods using DeepLab-V1~(VGG-16)~\cite{DEEPLAB14Chen,DEEPLAB18Chen} on the PASCAL VOC 2012~\cite{VOC10Ever} dataset in terms of mIoU. Numbers in bold indicate the best performance and underscored ones are the second best.}}
\vspace{-0.3cm}
\label{table:quantitative-vgg}
\begin{tabular}{c c c}
\specialrule{.1em}{.05em}{.05em}
\multicolumn{1}{l}{Method} & $val$ & $test$
\\ 
\hline
\hline
\multicolumn{3}{l}{\textit{Supervision}: Image-level labels~(10K) with Saliency~(3K)}
\\
\multicolumn{1}{l}{SeeNet$_{\text{NIPS'18}}$~\cite{SeeNet18Hou}} & 61.1 & 60.7
\\
\multicolumn{1}{l}{FickleNet$_{\text{CVPR'19}}$~\cite{FICKLE19Lee}} & 61.2 & 61.9
\\
\multicolumn{1}{l}{OAA$_{\text{ICCV'19}}$~\cite{OAA19Jiang}} & 63.1 & 62.8
\\
\multicolumn{1}{l}{ICD$_{\text{CVPR'20}}$~\cite{ICD20Fan}} & 64.0 & 63.9
\\
\hline
\multicolumn{2}{l}{\textit{Supervision}: Boxes~(10K)}
\\
\multicolumn{1}{l}{BoxSup$_{\text{ICCV'15}}$~\cite{BOXSUP15Dai}} & 62.0 & 64.6
\\
\multicolumn{1}{l}{WSSL$_{\text{ICCV'15}}$~\cite{WSSL15Pap}} & 60.6 & 62.2
\\
\multicolumn{1}{l}{SDI$_{\text{CVPR'17}}$~\cite{SDI17Kho}} & 65.7 & \underline{67.5}
\\
\multicolumn{1}{l}{BCM$_{\text{CVPR'19}}$~\cite{BOXDRIVEN19Song}} & 66.8 & -
\\
\multicolumn{1}{l}{Ours}
\\
\multicolumn{1}{l}{~~~~w/~$Y_\text{crf}$} & \underline{67.8} & -
\\
\multicolumn{1}{l}{~~~~w/~$Y_\text{ret}$} & 66.1 & -
\\
\multicolumn{1}{l}{~~~~w/~NAL} & \textbf{68.1} & \textbf{69.4}
\\
\hline
\multicolumn{2}{l}{\textit{Supervision}: Boxes~(9K) with Masks~(1K)}
\\
\multicolumn{1}{l}{BoxSup$_{\text{ICCV'15}}$~\cite{BOXSUP15Dai}} & 63.5 & 66.2
\\
\multicolumn{1}{l}{WSSL$_{\text{ICCV'15}}$~\cite{WSSL15Pap}} & 65.1 & 66.6
\\
\multicolumn{1}{l}{SDI$_{\text{CVPR'17}}$~\cite{SDI17Kho}} & 65.8 & \underline{66.9}
\\
\multicolumn{1}{l}{BCM$_{\text{CVPR'19}}$~\cite{BOXDRIVEN19Song}} & \underline{67.5} & -
\\
\multicolumn{1}{l}{Ours w/~NAL} & \textbf{70.5} & \textbf{71.5}
\\
\specialrule{.1em}{.05em}{.05em}
\end{tabular}
\vspace{0.3cm}
\end{table}

\begin{table}[t]
\centering
\footnotesize
\vspace{-.1cm}
\caption{\small{Quantitative comparison with state-of-the-art methods using DeepLab-V2~(ResNet-101)~\cite{DEEPLAB18Chen} on the PASCAL VOC 2012~\cite{VOC10Ever} dataset in terms of mIoU. Numbers in bold indicate the best performance and underscored ones are the second best. Contrary to others, Box2Seg~\cite{BTS20Kul} adopts UPerNet~\cite{UPer18Xiao} that consists of the FPN~\cite{FPN17Lin}, the PPM~\cite{PPM17Zhao}, and three decoders. $^\dagger$: models using pre-trained weights on MS-COCO~\cite{COCO14Lin}. $^*$: models using 10\% of the total images with ground-truth segmentation labels.}}
\vspace{-0.3cm}
\label{table:quantitative-res}
\begin{tabular}{c c c}
\specialrule{.1em}{.05em}{.05em}
\multicolumn{1}{l}{Method} & $val$ & $test$
\\ 
\hline
\hline
\multicolumn{3}{l}{\textit{Supervision}: Image-level labels~(10K) with Saliency~(3K)}
\\
\multicolumn{1}{l}{SeeNet$_{\text{NIPS'18}}$~\cite{SeeNet18Hou}} & 63.1 & 62.8
\\
\multicolumn{1}{l}{FickleNet$_{\text{CVPR'19}}$~\cite{FICKLE19Lee}} & 64.9 & 65.3
\\
\multicolumn{1}{l}{OAA$_{\text{ICCV'19}}$~\cite{OAA19Jiang}} & 65.2 & 66.4
\\
\multicolumn{1}{l}{ICD$_{\text{CVPR'20}}$~\cite{ICD20Fan}} & 67.8 & \underline{68.0}
\\
\hline
\multicolumn{2}{l}{\textit{Supervision}: Boxes~(10K)}
\\
\multicolumn{1}{l}{SDI$^\dagger_{\text{CVPR'17}}$~\cite{SDI17Kho}} & 74.2 & -
\\
\multicolumn{1}{l}{BCM$^\dagger_{\text{CVPR'19}}$~\cite{BOXDRIVEN19Song}} & 70.2 & -
\\
\multicolumn{1}{l}{Box2Seg$_{\text{ECCV'20}}$~\cite{BTS20Kul}} & \textbf{76.4} & -
\\
\multicolumn{1}{l}{Ours$^\dagger$}
\\
\multicolumn{1}{l}{~~~~w/~$Y_\text{crf}$} & 74.0 & -
\\
\multicolumn{1}{l}{~~~~w/~$Y_\text{ret}$} & 72.4 & -
\\
\multicolumn{1}{l}{~~~~w/~NAL} & \underline{74.6} & \textbf{76.1}
\\
\hline
\multicolumn{2}{l}{\textit{Supervision}: Boxes~(9K) with Masks~(1K)}
\\
\multicolumn{1}{l}{BCM$^\dagger_{\text{CVPR'19}}$~\cite{BOXDRIVEN19Song}} & 71.6 & -
\\
\multicolumn{1}{l}{Box2Seg$^*_{\text{ECCV'20}}$~\cite{BTS20Kul}} & \textbf{83.1} & -
\\
\multicolumn{1}{l}{Ours$^\dagger$ w/~NAL} & \underline{78.7} & \textbf{79.4}
\\
\specialrule{.1em}{.05em}{.05em}
\end{tabular}
\vspace{-.92cm}
\end{table}

\begin{table}[b]
\centering
\footnotesize
\caption{\small{Quantitative comparison for instance segmentation on the MS-COCO~\cite{COCO14Lin}~$test$ set.}}
\vspace{-0.3cm}
\label{table:coco}
\newcolumntype{C}[1]{>{\centering\arraybackslash}p{#1}}
\begin{tabular}{c C{0.45cm} C{0.45cm} C{0.45cm} C{0.45cm} C{0.45cm} C{0.45cm}}
\specialrule{.1em}{.05em}{.05em}
\multicolumn{1}{l}{Method} & $\text{AP}$ & $\text{AP}_{50}$ & $\text{AP}_{75}$ & $\text{AP}_S$ & $\text{AP}_M$ & $\text{AP}_L$
\\ 
\hline
\hline
\multicolumn{1}{l}{Mask-RCNN~\cite{MASKRCNN17He}} & 35.7 & 58.0 & 37.8 & 15.5 & 38.1 & 52.4
\\
\multicolumn{1}{l}{AISI~\cite{ASSOCIATE18Fan}} & 13.7 & 25.5 & 13.5 & 0.7 & 15.7 & 26.1
\\
\multicolumn{1}{l}{Ours~(`VOC-to-COCO')} & 16.9 & 38.2 & 13.0 & 7.3 & 17.1 & 26.5
\\
\multicolumn{1}{l}{Ours~(`COCO-to-COCO')} & 22.2 & 47.1 & 18.7 & 11.2 & 22.1 & 31.4
\\
\specialrule{.1em}{.05em}{.05em}
\end{tabular}
\end{table}

\section{Conclusion}
We have presented a novel pooling method for WSSS, dubbed BAP, using a background prior, that discriminates foreground and background regions inside object bounding boxes. We have shown that our BAP allows to produce better pseudo ground-truth labels compared to the conventional GAP. We have proposed a NAL for training a segmentation network, making it less susceptible to incorrect pseudo labels. Finally, we have shown that our approach achieves state-of-the-art performance on PASCAL VOC and MS-COCO.

\noindent \textbf{Acknowledgments.}
This work was supported by the National Research Foundation of Korea (NRF) grant funded by the Korea government (MSIP) (NRF-2019R1A2C2084816).


{\small
\bibliographystyle{ieee_fullname}
\bibliography{egbib}
}

\clearpage


\section*{Supplementary material}

# Background-Aware Pooling and Noise-Aware Loss for Weakly-Supervised Semantic Segmentation
## Supplement

Youngmin Oh    Beomjun Kim    Bumsub Ham*
School of Electrical and Electronic Engineering, Yonsei University

We present a detailed analysis for hyperparameters (Sec. 1), and provide a quantitative comparison between GAP and BAP (Sec. 2). We also show runtime comparisons of our pseudo label generator and other methods (Sec. 3). We then compare dot product based and cosine similarity based classifiers for semantic segmentation (Sec. 4), and describe training losses in detail (Sec. 5). We finally show more quantitative and qualitative results on PASCAL VOC 2012 [3] and MS-COCO [9] (Sec. 6).

## 1. Hyperparameters

**Grid size ($N$).** To analyze an effect of a grid size, we use different grid sizes for training a classification network using BAP and for generating pseudo segmentation labels (*i.e.*, $u_0$). Table 1(a) compares performance of pseudo labels on the PASCAL VOC 2012 [3] *train* set. We can see that training with a single query (*i.e.*, $N = 1$) provides the worst result. A plausible explanation is that a definite background might contain inhomogeneous clutter, and thus a larger grid size can help the classification network to use diverse queries, acting as a regularizer. On the other hand, for generating pseudo labels, we empirically find that a smaller grid size tends to give better results. Accordingly, we set the grid size $N$ to 4 for training the classification network, and to 1 for generating pseudo labels. We also show in Table 1(b) segmentation results, trained with corresponding pseudo labels in Table 1(a), on the PASCAL VOC 2012 *val* set. To this end, we use DeepLab-V1 [1, 2] with the standard cross-entropy loss. We can see that more accurate pseudo labels give better segmentation results.

**Damping and balance parameters ($\gamma$ and $\lambda$).** We show in Fig. 1(a) an effect of damping parameters on confidence scores $\sigma(\mathbf{p}) = (\alpha(\mathbf{p}))^\gamma$, where

$$\alpha(\mathbf{p}) = \frac{D_{c^*}(\mathbf{p})}{\max_c(D_c(\mathbf{p}))}. \quad (1)$$

Note that confidence scores turn into binary ones with a large value of $\gamma$, similar to BCM [15]. We show in Fig. 1(b)

*Corresponding author.

Table 1: (a) Comparison of our pseudo labels $Y_{\text{crf}}$ using different grid sizes on the PASCAL VOC 2012 [3] *train* set in terms of mIoU. (b) Comparison of mIoU scores using DeepLab-V1 [1, 2], trained with corresponding pseudo labels in (a), on the PASCAL VOC 2012 *val* set. The best performance is reported in bold and the second best is underlined.

| Grid Size $N$ | For Generating | | | Grid Size $N$ | For Generating | | |
|---|---|---|---|---|---|---|---|
| | 1 | 2 | 3 | | 1 | 2 | 3 |
| For Training 1 | 75.26 | 75.28 | 75.24 | For Training 1 | 64.88 | 64.81 | 64.83 |
| 2 | 76.68 | 76.63 | 76.57 | 2 | 65.90 | 65.88 | 65.87 |
| 3 | 78.36 | 78.24 | 78.18 | 3 | 67.66 | 67.67 | 67.69 |
| 4 | **78.70** | <u>78.50</u> | 78.48 | 4 | **67.82** | <u>67.79</u> | 67.74 |
| 5 | 78.30 | 78.10 | 78.12 | 5 | 67.73 | 67.65 | 67.66 |
| (a) | | | | (b) | | | |

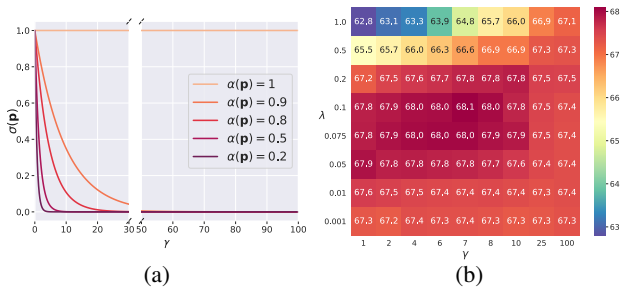

Figure 1: (a) Confidence scores $\sigma$ with different damping parameters $\gamma$. (b) Comparison of mIoU scores for different hyperparameters ($\gamma$ and $\lambda$) on the PASCAL VOC 2012 [3] *val* set. Best viewed in color.

the performance of DeepLab-V1 [1, 2] for different parameters $\gamma$ and $\lambda$ on the PASCAL VOC 2012 [3] *val* set. We select a pair of parameters which provides the best result (*i.e.*, $\gamma = 7$, $\lambda = 0.1$).

## 2. Comparison of GAP and BAP

To further verify that BAP gives better CAMs [17] than GAP, we quantify the quality of CAMs for different pooling methods. To this end, we consider normalized CAM scores $u_c$ (Eq. (5) of the main paper) as *True-*

Table 2: Quantitative comparison of CAMs using different pooling methods on the PASCAL VOC 2012 [3] *train* set.

| Method | aero | bike | bird | boat | bot | bus | car | cat | cha | cow | tab | dog | hor | mbik | pers | plnt | she | sofa | trai | tv | mean |
|---|---|---|---|---|---|---|---|---|---|---|---|---|---|---|---|---|---|---|---|---|---|
| GAP | 62.7 | 27.3 | 68.4 | 59.1 | 86.2 | 91.4 | 86.7 | 83.3 | 51.5 | 82.6 | 74.6 | 81.6 | 74.7 | 69.7 | 67.2 | 68.2 | 79.0 | 69.5 | 84.1 | 92.1 | 73.0 |
| BAP | 69.9 | 68.2 | 71.9 | 70.7 | 84.2 | 93.5 | 87.5 | 88.3 | 57.2 | 84.7 | 66.4 | 85.0 | 75.8 | 79.1 | 71.5 | 79.0 | 83.1 | 66.0 | 88.8 | 87.1 | 77.9 |

Table 3: Comparison of mIoU scores on the PASCAL VOC 2012 [3] *val* set.

| Method | val |
|---|---|
| *Architecture*: DeepLab-V1 [1, 2] | |
| SDI [7] | 65.7 |
| BCM [15] | 66.8 |
| Ours w/ NAL | |
| GAP | 66.3 |
| BAP | 68.1 |
| *Architecture*: DeepLab-V2 [2] | |
| SDI [7] | 74.2 |
| BCM [15] | 70.2 |
| Ours w/ NAL | |
| GAP | 72.6 |
| BAP | 74.6 |

Table 4: Runtime comparison of our pseudo label generator and other segmentation methods. We report the average runtime on the PASCAL VOC 2012 [3] *val* set.

| Times (s) | GrabCut [14] | MCG [11] | WSSL [10] | Ours |
|---|---|---|---|---|
| CPU | 1.9 | 25.5 | 0.4 | 0.4 |
| GPU | - | - | - | 0.1 |
| Total | 1.9 | 25.5 | 0.4 | 0.5 |

Table 5: Comparison of mIoU scores using DeepLab-V1 [1, 2] on the PASCALS VOC 2012 [3] *val* set. For our NAL, we use two different classifiers. DP: a dot product based classifier. CS: a cosine similarity based classifier.

| Method | val |
|---|---|
| SDI [7] | 65.7 |
| BCM [15] | 66.8 |
| Ours w/ NAL | |
| using DP | 67.8 |
| using CS ($\tau = 20$) | 68.1 |

Table 6: Results of DeepLab-V2 with different initialization on the PASCAL VOC 2012 [3] *val* set. Numbers in brackets indicate fully-supervised performance. FS denotes performance relative to the fully-supervised one.

| Method | w/ ImageNet | | w/ MS-COCO | |
|---|---|---|---|---|
| | mIoU | FS | mIoU | FS |
| SDI [7] | 69.4 (74.5) | 93.2 | 74.2 (77.7) | 95.5 |
| BCM [15] | - | - | 70.2 (74.5) | 94.2 |
| Ours w/ NAL | 70.9 (73.4) | 96.6 | 74.6 (77.5) | 96.3 |

*Positives* (TP$_c$) if they are activated inside ground-truth regions for a class $c$. Otherwise, we regard them as *False-Positives* (FP$_c$). We define the precision for the class $c$ as: $\frac{\text{TP}_c}{\text{TP}_c + \text{FP}_c}$. We compare in Table 2 the per-class precision of CAMs obtained with different pooling methods. We can clearly see that BAP outperforms GAP, especially for *bike*, *boat*, and *plant* classes. A plausible reason is that GAP may hinder classifier weights since it aggregates a mixture of foreground and background features inside object bounding boxes. BAP overcomes this problem by computing a background attention map adaptively, encouraging the classifier weights to focus more on foreground features.

We also evaluate the classification networks using different pooling methods on the PASCAL VOC 2012 [3] *val* set. To this end, we consider ground-truth boxes as pre-obtained ones to crop each object. We find that the classification accuracy increases from 80.3% to 82.3% by using BAP instead of GAP. This suggests that BAP could be helpful in improving object classification/detection performance.

Table 3 shows mIoU scores of DeepLab-V1 and -V2 [1, 2], trained with different pseudo labels using GAP or BAP, on the PASCAL VOC 2012 *val* set. We can see that the models trained with the pseudo labels using GAP ('GAP') achieve competitive mIoU scores to BCM [15], demonstrating once again the effectiveness of our approach to using a classification network with bounding boxes.

## 3. Runtime

We show in Table 4 a runtime comparison for generating pseudo labels. For fair comparison, we use a NVIDIA Titan RTX GPU with an Intel i5 3.7GHz for all experiments. For the result of MCG [11], we use a publicly available MATLAB implementation, and choose the best segment proposal for each bounding box. Our pseudo label generator mainly consists of two parts: (1) feature extraction with computing a unary term (*i.e.*, $u_c$ and $u_0$); (2) DenseCRF [8]. The first part takes 0.1 seconds on the GPU and the second one takes 0.4 seconds on the CPU. WSSL also adopts DenseCRF to generate pseudo segmentation labels from object bounding boxes. Compared to WSSL, our pseudo label generator requires a negligible overhead (0.1 seconds), while it brings substantial mIoU gains. For example, our pseudo labels $Y_{\text{crf}}$ achieve the mIoU gain of 9.0/8.1% over WSSL on the PASCAL VOC 2012 [3] *train*/*val* sets (see Table 1 in the main paper).

## 4. Classifier for semantic segmentation

Here we provide a detailed description of two different classifiers for semantic segmentation. First, a dot product (DP) based classifier computes a probability for a class $c$ as follows:

$$H_c(\mathbf{p}) = \frac{e^{\phi(\mathbf{p}) \cdot W_c}}{\sum_i e^{\phi(\mathbf{p}) \cdot W_i}}. \tag{2}$$

Table 7: Per-class results of DeepLab-V1 [1, 2] in terms of mIoU on the PASCAL VOC 2012 [3] dataset: (a) the results on the *val* set, (b) the results on the *test* set. All numbers except for ours are taken from corresponding papers.

| Method | bkg | aero | bike | bird | boat | bot | bus | car | cat | cha | cow | tab | dog | hor | mbik | pers | plnt | she | sofa | trai | tv | mean |
|---|---|---|---|---|---|---|---|---|---|---|---|---|---|---|---|---|---|---|---|---|---|---|
| SDI [7] | - | - | - | - | - | - | - | - | - | - | - | - | - | - | - | - | - | - | - | - | - | 65.7 |
| BCM [15] | 89.8 | 68.3 | 27.1 | 73.7 | 56.4 | 72.6 | 84.2 | 75.6 | 79.9 | 35.2 | 78.3 | 53.2 | 77.6 | 66.4 | 68.1 | 73.1 | 56.8 | 80.1 | 45.1 | 74.7 | 54.6 | 66.8 |
| Ours w/ NAL | 91.0 | 78.1 | 30.8 | 83.1 | 62.4 | 67.7 | 85.9 | 78.4 | 83.2 | 34.8 | 77.3 | 52.7 | 76.2 | 73.6 | 70.3 | 74.0 | 49.9 | 77.7 | 40.5 | 79.3 | 64.6 | 68.1 |

(a)

| Method | bkg | aero | bike | bird | boat | bot | bus | car | cat | cha | cow | tab | dog | hor | mbik | pers | plnt | she | sofa | trai | tv | mean |
|---|---|---|---|---|---|---|---|---|---|---|---|---|---|---|---|---|---|---|---|---|---|---|
| SDI [7] | - | 78.1 | 31.1 | 72.4 | 61.0 | 67.2 | 84.2 | 78.2 | 81.7 | 27.6 | 68.5 | 62.1 | 76.9 | 70.8 | 78.2 | 76.3 | 51.7 | 78.3 | 48.3 | 74.2 | 58.6 | 67.5 |
| Ours w/ NAL | 91.7 | 80.5 | 32.5 | 80.5 | 61.1 | 66.2 | 86.6 | 79.2 | 85.4 | 28.6 | 73.6 | 60.8 | 79.4 | 71.5 | 78.2 | 73.9 | 57.5 | 81.1 | 48.2 | 78.7 | 62.8 | 69.4 |

(b)

Table 8: Per-class results of DeepLab-V2 [2] in terms of mIoU on the PASCAL VOC 2012 [3] dataset: (a) the results on the *val* set, (b) the results on the *test* set. All numbers except for ours are taken from corresponding papers.

| Method | bkg | aero | bike | bird | boat | bot | bus | car | cat | cha | cow | tab | dog | hor | mbik | pers | plnt | she | sofa | trai | tv | mean |
|---|---|---|---|---|---|---|---|---|---|---|---|---|---|---|---|---|---|---|---|---|---|---|
| SDI [7] | - | - | - | - | - | - | - | - | - | - | - | - | - | - | - | - | - | - | - | - | - | 74.2 |
| BCM [15] | - | - | - | - | - | - | - | - | - | - | - | - | - | - | - | - | - | - | - | - | - | 70.2 |
| Ours w/ NAL | 93.1 | 83.8 | 33.1 | 87.4 | 64.1 | 78.3 | 93.2 | 84.4 | 89.3 | 39.3 | 84.2 | 61.9 | 85.2 | 80.1 | 76.9 | 79.4 | 57.5 | 84.6 | 51.9 | 85.8 | 74.7 | 74.6 |

(a)

| Method | bkg | aero | bike | bird | boat | bot | bus | car | cat | cha | cow | tab | dog | hor | mbik | pers | plnt | she | sofa | trai | tv | mean |
|---|---|---|---|---|---|---|---|---|---|---|---|---|---|---|---|---|---|---|---|---|---|---|
| Ours w/ NAL | 93.6 | 85.0 | 34.5 | 90.4 | 65.2 | 75.9 | 92.0 | 85.2 | 91.3 | 38.4 | 84.9 | 67.6 | 86.6 | 85.8 | 82.3 | 79.3 | 62.3 | 87.7 | 60.0 | 81.8 | 69.2 | 76.1 |

(b)

We omit bias terms to make the classifier weight $W$ to be more representative for corresponding classes. The classifier weights depend on both magnitude and direction of the features $\phi$, and thus this classifier is not robust to intra-class variations. It has recently been shown that a cosine similarity (CS) based classifier could outperform the DP based one [4, 12, 16]. Specifically, the CS based classifier modifies Eq. (2) as follows:

$$H_c(\mathbf{p}) = \frac{e^{\tau \frac{\phi(\mathbf{p})}{\|\phi(\mathbf{p})\|} \cdot \frac{W_c}{\|W_c\|}}}{\sum_i e^{\tau \frac{\phi(\mathbf{p})}{\|\phi(\mathbf{p})\|} \cdot \frac{W_i}{\|W_i\|}}}, \quad (3)$$

where we denote by $\tau$ a scale parameter. This encourages classifier weights and features to lie on a hypersphere, whose radius is $\tau$. Since this classifier considers the angle between the feature $\phi$ and the classifier $W_i$ only, we expect that it is more robust against intra-class variations than the DP based one.

We report in Table 5 mIoU scores of DeepLab-V1 [1, 2] using DP and CS based classifiers on the PASCAL VOC 2012 [3] *val* set. We can see that the CS based classifier outperforms the DP based one slightly. Accordingly, we exploit the CS based classifier with the scale parameter $\tau$ of 20. It is worth noting that our model using the DP based classifier still outperforms the state of the art [7, 15].

## 5. Training losses

We describe more details on training losses to deal with the unreliable regions $\sim\mathcal{S}$ (see Table 3 in the main paper).

We define entropy regularization [5] as follows:

$$\mathcal{L}_{ER} = -\frac{1}{|\sim\mathcal{S}|} \sum_{\mathbf{p}\in\sim\mathcal{S}} \sum_c H_c(\mathbf{p}) \log H_c(\mathbf{p}), \quad (4)$$

where $|\cdot|$ indicates the number of pixels. This encourages a $(L+1)$-dimensional probability map $H$ to have one clear peak at each position $\mathbf{p}$ for a confident prediction. The overall loss is defined as follows:

$$\mathcal{L} = \mathcal{L}_{ce} + \lambda \mathcal{L}_{ER}, \quad (5)$$

where we set $\lambda$ to 0.1. Motivated by the work of [13], we also use a bootstrapping technique as follows:

$$\mathcal{L}_{BS} = -\frac{1}{|\sim\mathcal{S}|} \sum_{\mathbf{p}\in\sim\mathcal{S}} \sum_c Y_c(\mathbf{p}) \log H_c(\mathbf{p}), \quad (6)$$

where $Y_c$ indicates the $c$-th element of a new target $Y$. Specifically, we define the new target vector at each position $\mathbf{p}$ as follows:

$$Y(\mathbf{p}) = \beta y(\mathbf{p}) + (1-\beta) H(\mathbf{p}), \quad (7)$$

where $\beta$ is a balance parameter between pseudo ground-truth labels $Y_{\text{crf}}$ and model predictions $H$. $y$ is a $(L+1)$-dimensional one-hot vector, with a value of 1 at the $c$-th dimension, where $c = Y_{\text{crf}}(\mathbf{p})$. The bootstrapping technique is similar to an EM-like algorithm, where the new target vector is estimated in the E-step, and the model is optimized to better predict such a new target in the M-step. We adjust the balance parameter $\beta$ from 1 to 0 by using the poly

schedule during training. The overall loss is then defined as follows:

$$\mathcal{L} = \mathcal{L}_{ce} + \lambda \mathcal{L}_{BS}, \tag{8}$$

where we also set $\lambda$ to 0.1.

## 6. More segmentation results

**PASCAL VOC.** To test fully-supervised performance, we train DeepLab-V1 [1, 2] with ground-truth segmentation labels, achieving a mIoU score of 69.5% on the PASCAL VOC 2012 [3] *val* set. This is similar to the mIoU scores of 69.1% and 69.8% reported in [7, 15], respectively. We show in Table 6 mIoU scores of DeepLab-V2 [2] using different initialization. For comparison, we also provide fully-supervised performance. We can see that our approach gives better results than other methods.

Table 7 compares per-class mIoU scores of DeepLab-V1 on the PASCAL VOC 2012 dataset. We can clearly see that our approach outperforms other WSSS methods by a large margin on both *val* and *test* sets. For example, our approach gives better results than BCM and SDI [7] for 13 and 14 classes, respectively. This demonstrates the importance of high-quality pseudo labels. We also provide in Table 8 per-class mIoU scores of DeepLab-V2 on the same dataset. Compared to the results of Table 7, we can see that using a deeper CNN improves the mIoU performance for all classes, achieving a new state of the art.

We show in Fig. 2 visual examples of $u_c$, $u_o$, and pseudo segmentation labels. We can see that our approach using BAP generates high-quality background attention maps $u_o$, leading to better CAMs $u_c$. Also, we show that $Y_{\text{ret}}$ is effective to identify unreliable regions. For example, the *person*'s legs in the fourth row are incorrectly labeled as a *cow* class, which can be marked by our approach. We show in Figs. 3 and 4 qualitative results of DeepLab-V1 and -V2 on the PASCAL VOC 2012 dataset, respectively. We can see that the networks trained with pseudo segmentation labels only (Weak) already yield satisfactory results, and exploiting a small number of ground-truth pixel-level labels (Semi) provides more accurate results with sharp object boundaries. The last rows for each figure show failure examples.

**MS-COCO.** We show in Table 9 AP scores of Mask-RCNN [6] on the MS-COCO [9] *test* set. We train Mask-RCNN with each of our pseudo labels, *i.e.*, $Y_{\text{crf}}$ and $Y_{\text{ret}}$. The behavior of AP scores is similar to that of mIoU scores for semantic segmentation (see Tables 4 and 5 in the main paper). Compared to the results of Table 6 in the main paper, we find that exploiting both $Y_{\text{crf}}$ and $Y_{\text{ret}}$ gives better results especially for AP, $AP_{50}$, $AP_{75}$, $AP_S$, and $AP_M$. This again confirms two pseudo labels $Y_{\text{crf}}$ and $Y_{\text{ret}}$ are complementary to each other.

In Fig. 5, we show visual examples of pseudo segmen-

Table 9: Quantitative comparison of Mask-RCNN [6], trained with either $Y_{\text{crf}}$ or $Y_{\text{ret}}$, on the MS-COCO [9] *test* set.

| Method | AP | $AP_{50}$ | $AP_{75}$ | $AP_S$ | $AP_M$ | $AP_L$ |
|---|---|---|---|---|---|---|
| VOC-to-COCO | | | | | | |
| BAP: $Y_{\text{crf}}$ | 17.0 | 35.6 | 14.4 | 5.8 | 16.6 | 28.4 |
| BAP: $Y_{\text{ret}}$ | 12.7 | 35.1 | 6.5 | 6.1 | 12.5 | 19.2 |
| COCO-to-COCO | | | | | | |
| BAP: $Y_{\text{crf}}$ | 20.7 | 41.9 | 18.4 | 8.0 | 20.2 | 32.2 |
| BAP: $Y_{\text{ret}}$ | 17.5 | 43.5 | 11.5 | 8.9 | 16.7 | 24.9 |

tation labels on the MS-COCO *train* set. We find that pseudo labels of VOC-to-COCO show reasonable results even for unseen classes (*e.g.*, *giraffe* and *pizza* classes), demonstrating the effectiveness of our pseudo label generator. We also show in Fig. 6 instance segmentation results of Mask-RCNN on the MS-COCO dataset. We can see that the model trained with COCO-to-COCO gives better results than the one using VOC-to-COCO.

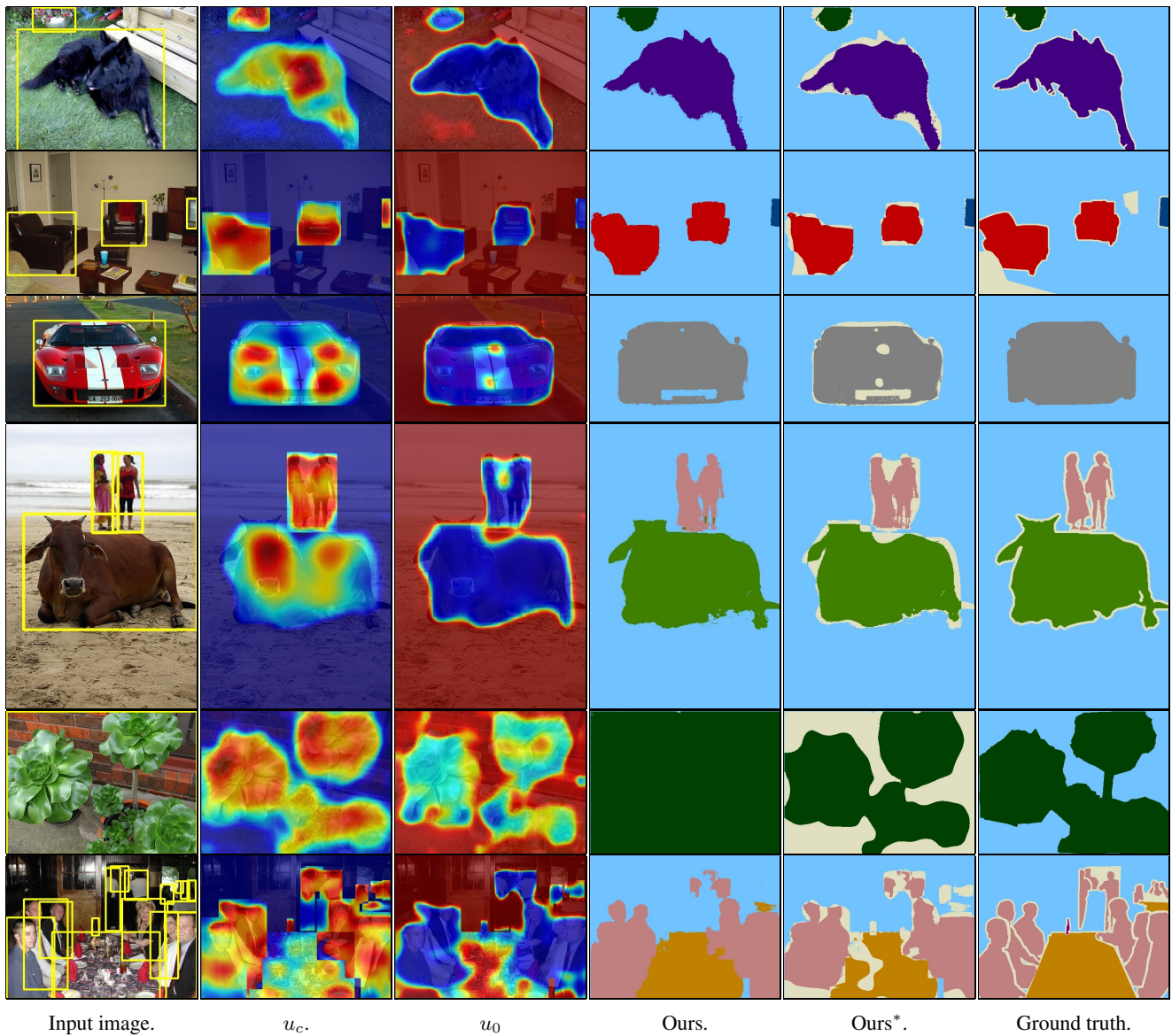

Figure 2: Visual examples of $u_c$, $u_0$, and corresponding pseudo segmentation labels on the PASCAL VOC 2012 [3] *train* set. We overlap $u_c$ over different classes for visualization. Ours: our pseudo labels $Y_{\text{crf}}$. Ours*: the same ones but with an indication of unreliable regions using $Y_{\text{ret}}$. Best viewed in color.

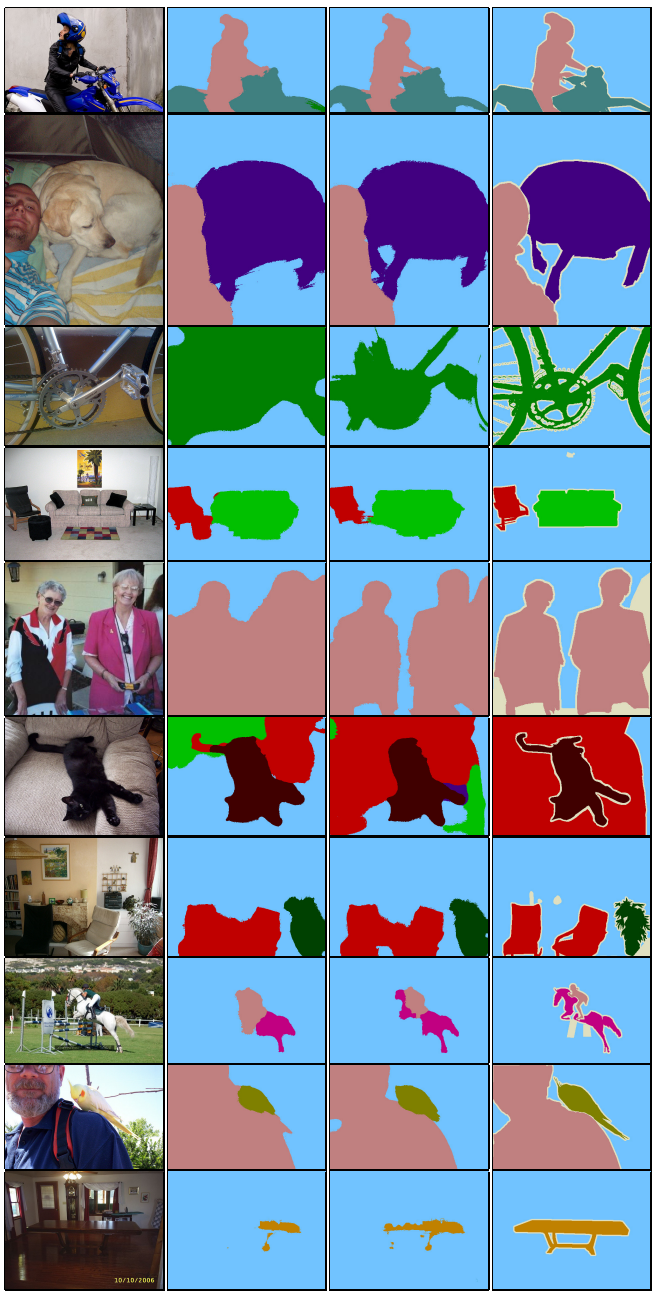
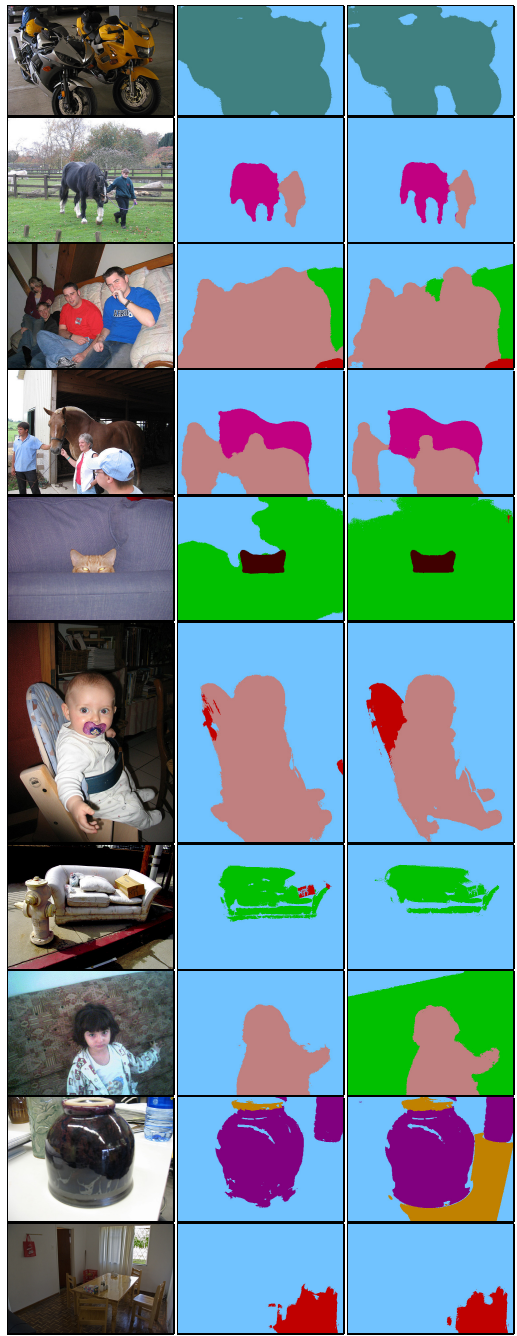

| Input image. | Ours (Weak). | Ours (Semi). | Ground truth. |

(a) Results on *val* set.

| Input image. | Ours (Weak). | Ours (Semi). |

(b) Results on *test* set.

Figure 3: Qualitative examples of our final model using DeepLab-V1 [1, 2] on the PASCAL VOC 2012 [3] dataset. Best viewed in color.

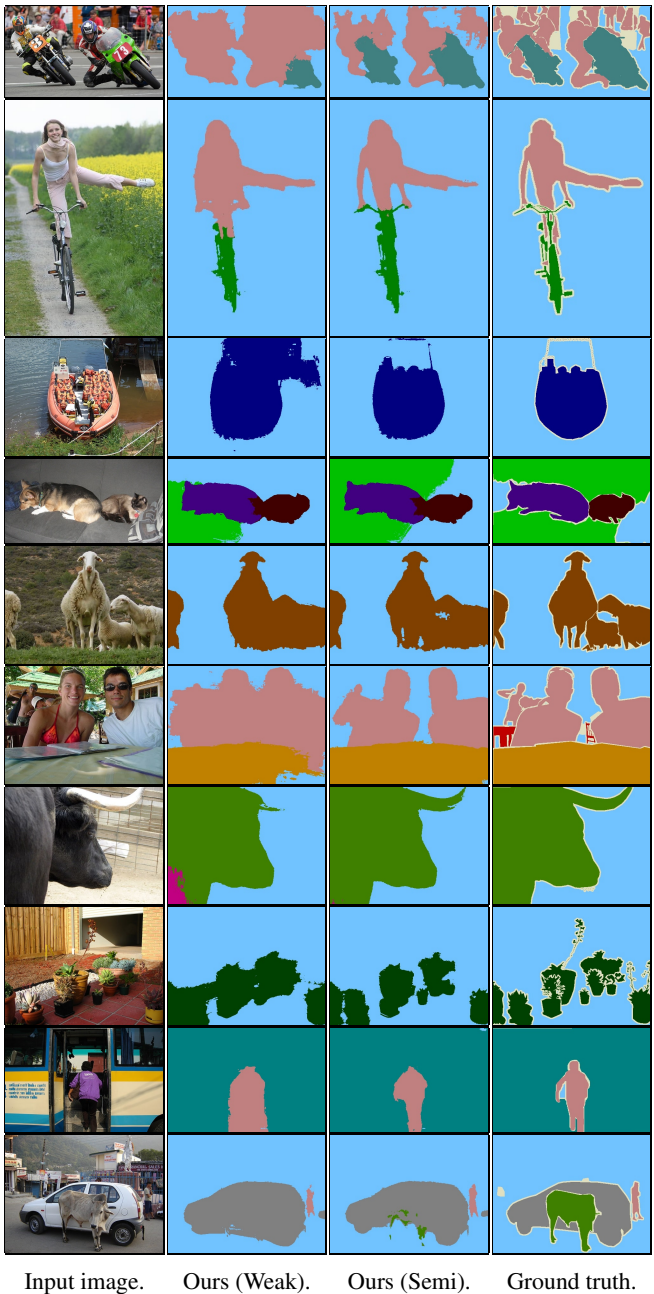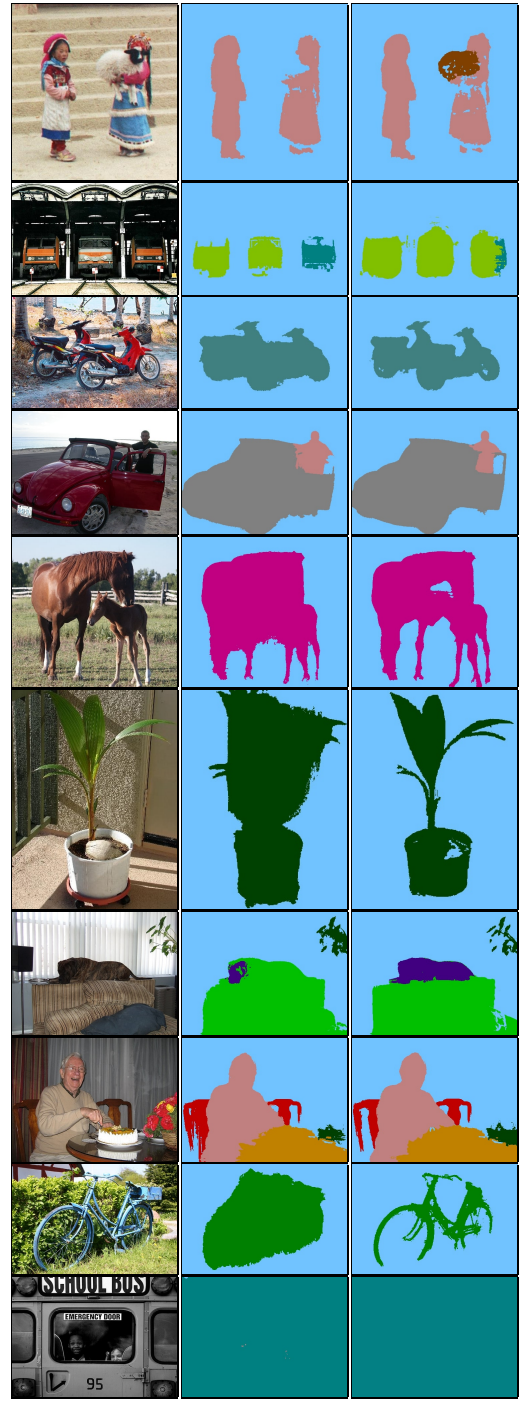

Figure 4: Qualitative examples of our final model using DeepLab-V2 [2] on the PASCAL VOC 2012 [3] dataset. Best viewed in color.

| $Y_{\text{crf}}.$ | $Y_{\text{ret}}.$ | $Y_{\text{crf}}.$ | $Y_{\text{ret}}.$ | Ground truth. |
|---|---|---|---|---|
| VOC-to-COCO | | COCO-to-COCO | | |

Figure 5: Visual examples of pseudo segmentation labels on the MS-COCO [9] *train* set. Note that VOC-to-COCO do not use any training samples of MS-COCO during training. Best viewed in color.

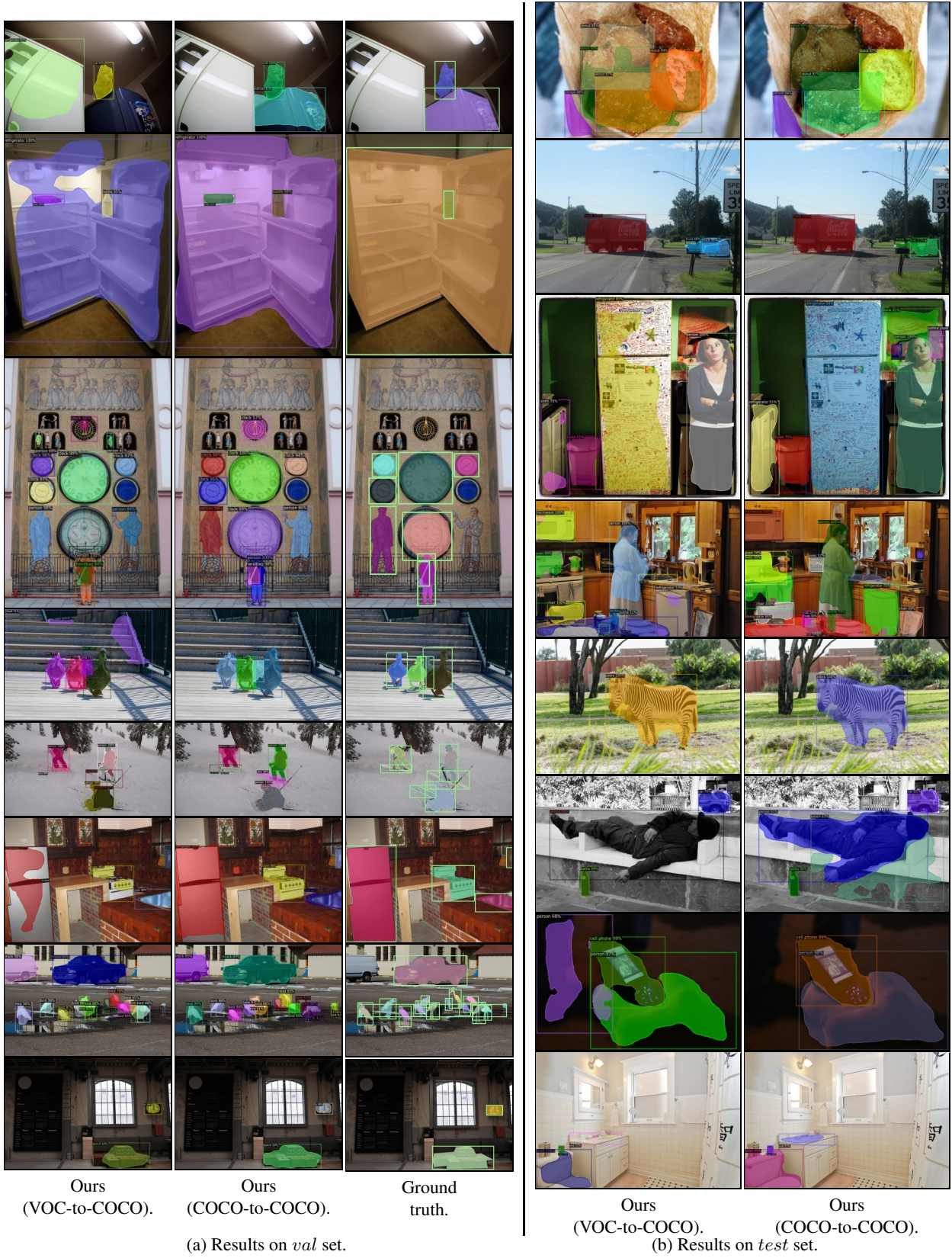

Figure 6: Qualitative results of Mask-RCNN [6] on the MS-COCO [9] dataset. We train Mask-RCNN with two pseudo labels $Y_{\text{crf}}$ and $Y_{\text{ret}}$, and compute the binary cross-entropy loss for the regions $\mathcal{S}$ only. Best viewed in color.



\end{document}